%% file: acl.tex
\pdfoutput=1

\documentclass[11pt]{article}
\usepackage{amsmath,amssymb}
\usepackage{acl}
\usepackage{multirow}
\usepackage{times}
\usepackage{latexsym}
\usepackage{booktabs}
\usepackage{graphicx}
\usepackage{subcaption}

\usepackage[T1]{fontenc}

\usepackage[utf8]{inputenc}

\usepackage{microtype}

\usepackage{graphicx}
\usepackage{svg}
\usepackage{enumitem}
\usepackage{lipsum}

\newcommand{\modelname}{\textbf{COGMEN}}

\usepackage{soul}

\title{COGMEN: COntextualized GNN based Multimodal Emotion recognitioN}

\author{Abhinav Joshi\qquad Ashwani Bhat \\  \textbf{Ayush Jain}\qquad  \textbf{Atin Vikram Singh} \qquad \textbf{Ashutosh Modi} \\
 Indian Institute of Technology Kanpur (IIT-K)\\
 \texttt{\{ajoshi, ashutoshm\}@cse.iitk.ac.in} \\
 \texttt{\{ashubhat44\}@gmail.com} \\ 
 \texttt{\{aayushj, atinvs\}@iitk.ac.in}
}

\begin{document}
\maketitle
\input{abstract}

\input{Introduction}
\input{RelatedWorks}

\input{Architecture}
\input{Experiments}

\input{Results-Analysis}

\input{Discussion}
\input{Conclusion}
\input{Acknowledgements}

\bibliography{references}
\bibliographystyle{acl_natbib}

\clearpage
\newpage
\appendix
\input{Appendix}

\end{document}

%% file: abstract.tex
\begin{abstract}
Emotions are an inherent part of human interactions, and consequently, it is imperative to develop AI systems that understand and recognize human emotions. During a conversation involving various people, a person's emotions are influenced by the other speaker's utterances and their own emotional state over the utterances. In this paper, we propose COntextualized Graph Neural Network based Multimodal Emotion recognitioN (COGMEN) system that leverages local information (i.e., inter/intra dependency between speakers) and global information (context). The proposed model uses Graph Neural Network (GNN) based architecture to model the complex dependencies (local and global information) in a conversation. Our model gives state-of-the-art (SOTA) results on IEMOCAP and MOSEI datasets, and detailed ablation experiments show the importance of modeling information at both levels. 
\end{abstract}

%% file: Introduction.tex
\section{Introduction}
Emotions are intrinsic to humans and guide their behavior and are indicative of the underlying thought process \cite{minsky2007emotion}. Consequently, understanding and recognizing emotions is vital for developing AI technologies (e.g., personal digital assistants) that interact directly with humans. During a conversation between a number of people, there is a constant ebb and flow of emotions experienced and expressed by each person. The task of multimodal emotion recognition addresses the problem of monitoring the emotions expressed (via various modalities, e.g., video (face), audio (speech)) by individuals in different settings such as conversations. 


\begin{figure}[t]
\centering
  \includegraphics[scale=0.18]{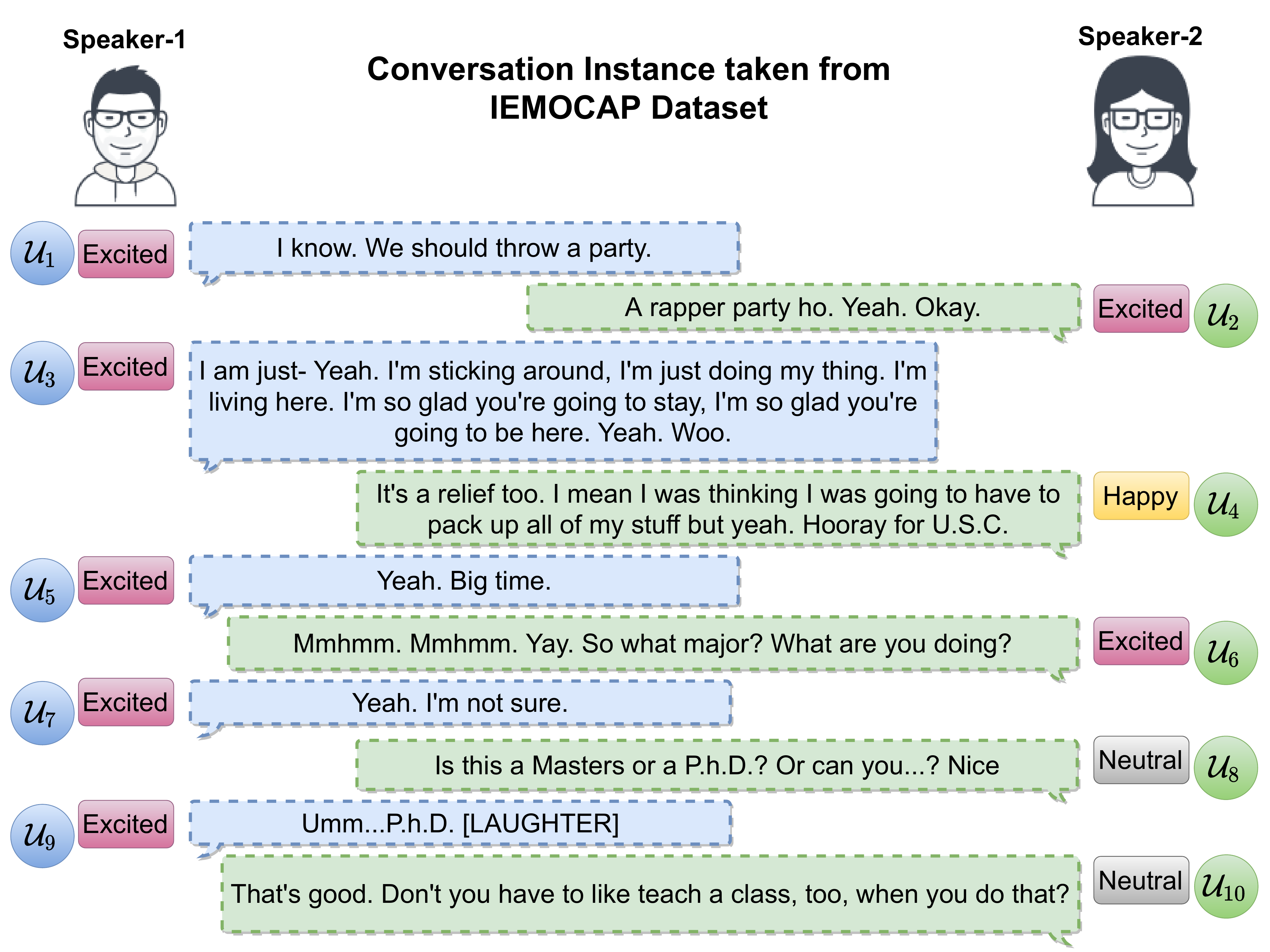}
  \caption{{An example conversation between two speakers, with corresponding emotions evoked for each utterance.}}
  \label{fig:conversation}
\end{figure}

Emotions are physiological, behavioral, and communicative reactions to cognitively processed stimuli \cite{planalp_fitness_fehr_2018}. Emotions are often a result of internal physiological changes, and these physiological reactions may not be noticeable by others and are therefore intra-personal. For example, in a conversational setting, an emotion may be a communicative reaction that has its origin in a sentence spoken by another person, acting as a stimulus. 
The emotional states expressed in utterances correlate with the context directly; for example, if the underlying context is about a happy topic like celebrating a festival or description of a vacation, there will be more positive emotions like joy and surprise. Consider the example shown in Figure \ref{fig:conversation}, where the context depicts an exciting conversation. Speaker-1 being excited about his admission affects the flow of emotions in the entire context. The emotion states of Speaker-2 show the dependency on Speaker-1 in $\mathcal{U}_2,\mathcal{U}_4$ and $\mathcal{U}_6$, and maintains intra-personal state depicted in $\mathcal{U}_8$ and $\mathcal{U}_{10}$ by being curious about the responses of Speaker-1. The example conversation portrays the effect of global information as well as inter and intra dependency of speakers on the emotional states of the utterances. Moreover, emotions are a multimodal phenomenon; a person takes cues from different modalities (e.g., audio, video) to infer the emotions of others, since, very often, the information in different modalities complement each other.  In this paper, we leverage these intuitions and propose \modelname: \textbf{CO}ntextualized \textbf{G}raph neural network based \textbf{M}ultimodal \textbf{E}motion recognitio\textbf{N} architecture  
that addresses both, the effects of context on the utterances and inter and intra dependency for predicting the per-utterance emotion of each speaker during the conversation. There has been a lot of work on unimodal (using text only) prediction, but our focus is on multimodal emotion prediction. As is done in literature on multimodal emotion prediction, we do not focus on comparison with unimodal models. As shown via experiments and ablation studies, our model leverages both the sources (i.e., local and global) of information to give state-of-the-art (SOTA) results on the multimodal emotion recognition datasets IEMOCAP and MOSEI. In a nutshell, we make the following contributions in this paper:
\begin{itemize}[noitemsep,topsep=0pt]
\item We propose a Contextualized Graph Neural Network (GNN) based Multimodal Emotion Recognition architecture for predicting per utterance per speaker emotion in a conversation. Our model leverages both local and global information in a conversation. We use GraphTransformers \cite{GraphTransformer} for modeling speaker relations in multimodal emotion recognition systems. 
\item Our model gives SOTA results on the multimodal Emotion recognition datasets of IEMOCAP and MOSEI. 
\item We perform a thorough analysis of the model and its different components to show the importance of local and global information along with the importance of the GNN component. We release the code for models and experiments: \url{https://github.com/Exploration-Lab/COGMEN} 
\end{itemize}

%% file: RelatedWorks.tex
\section{Related Work} \label{sec:relatedwork}

Emotion recognition is an actively researched problem in NLP \cite{ERC_survey_2021, multimodal_survey}. The broad applications ranging from emotion understanding systems,  opinion mining from a corpus to emotion generation have attracted active research interest in recent years \cite{dhuheir2021emotion_applicationHealthcare, franzen2021developing_applicationVideocon, vinola2015survey_application, kolakowska2014emotion_applicationSurvey,colombo-etal-2019-affect,sepehr-iva, goswamy-etal-2020-adapting, singh-etal-2021-end, harsh-shapes-2021,gargi-fine-grained-emotion}. Availability of benchmark multimodal datasets, such as CMU-MOSEI \cite{zadeh2018multimodal}, and IEMOCAP \cite{busso2008iemocap}, have accelerated the progress in the area. Broadly speaking, most of the existing work in this area can be categorized mainly into two areas: \textit{unimodal} approaches and \textit{multimodal} approaches. Unimodal approaches tend to consider the text as a prominent mode of communication and solve the emotion recognition task using only text modality. In contrast, multimodal approaches are more naturalistic and consider multiple modalities (audio+video+text) and fuse them to recognize emotions. In this paper, we propose a multimodal approach to emotion recognition. Nevertheless, we briefly outline some of the prominent unimodal approaches as some of the techniques are applicable to our setting. 

\noindent\textbf{Unimodal Approaches:} COSMIC \cite{yu2019deep_coattention} performs text only emotion classification problem by leveraging commonsense knowledge.  DialogXL \cite{shen2020dialogxl} uses XLnet \cite{yang2020xlnet} as architecture in dialogue feature extraction. CESTa \cite{wang2020contextualized_cesta} captures the emotional consistency in the utterances using Conditional Random Fields \cite{10.5555/645530.655813ConditionalRandomFields} for boosting the performance of emotion classification. Other popular approaches parallel to our work use graph-based neural networks as their baseline and solve the context propagation issues in RNN-based architectures, including DialogueGCN \cite{ghosal2019dialoguegcn}, RGAT \cite{ishiwatari2020relation_rgat}, ConGCN \cite{DBLP:conf/ijcai/ZhangWSLZZ19_ConGCN}, and SumAggGin \cite{sheng2020summarize_sumaggin}. Some of the recent approaches like DAG-ERC \cite{shen2021directed_dagerc} combine the strengths of conventional graph-based neural models and recurrence-based neural models.


\noindent\textbf{Multimodal Approaches:} Due to the high correlation between emotion and facial cues \cite{ekman1993facial}, fusing modalities to improve emotion recognition has drawn considerable interest \cite{multimodal_survey}. Some of the initial approaches include  \citet{Datcu2015SemanticAD}, who fused acoustic information with visual cues for emotion recognition. \citet{wollmer2010context} use  contextual information for emotion recognition in a multimodal setting. In the past decade, the growth of deep learning has motivated a wide range of approaches in multimodal settings. The Memory Fusion network (MFN) \cite{zadeh2018memory_MFN} proposes synchronizing multimodal sequences using multi-view gated memory storing intra-view and cross-view interactions through time. Graph-MFN \cite{bagher-zadeh-etal-2018-multimodal_GraphMFN} extends the idea of MFN and introduces Dynamic Fusion Graph (DFG), which learns to model the n-modal interactions and alter its structure dynamically to choose a fusion graph based on the importance of each n-modal dynamics during inference. Conversational memory network (CMN) \cite{hazarika-etal-2018-conversational_CMN} leverages contextual information from the conversation history and uses gated recurrent units to model past utterances of each speaker into memories. Tensor fusion Network (TFN) \cite{zadeh2017tensor_TFN} uses an outer product of the modalities. Other popular approaches include DialogueRNN \cite{majumder2019dialoguernn} that proposes an attention mechanism over the different utterances and models emotional dynamics by its party GRU and global GRU. B2+B4 \cite{kumar2020gated_b2b4}, use a conditional gating mechanism to learn cross-modal information. bc-LSTM \cite{poria2017context_bclstm} proposes an LSTM-based model that captures contextual information from the surrounding utterances. Multilogue-Net \cite{shenoy2020multilogue} proposes a solution based on a context-aware RNN and uses pairwise attention as a fusion mechanism for all three modalities (audio, video, and text). Recently, \citet{delbrouck2020transformer_tbje} proposed TBJE, a transformer-based architecture with modular co-attention \cite{yu2019deep_coattention} to encode multiple modalities jointly. CONSK-GCN \cite{9428438_CONSK_GCN} uses graph convolutional network (GCN) with knowledge graphs. \citet{lian2020conversational} use GNN based architecture for Emotion Recognition using text and speech modalities. Af-CAN \cite{afCAN} proposes RNN based on contextual attention for modeling the transaction and dependence between speakers.

%% file: Architecture.tex
\section{Proposed Model}


In a conversation involving different speakers, there is a continuous ebb and flow in the emotions of each of the speakers, usually triggered by the context and reactions of other speakers. Inspired by this intuition, we propose a multimodal emotion prediction model that leverages contextual information, inter-speaker and intra-speaker relations in a conversation.   

In our model, we leverage both the context of dialogue and the effect of nearby utterances. We model these two sources of information via two means: \textbf{1) Global Information}: How to capture the impact of underlying context on the emotional state of an utterance? \textbf{2) Local information}: How to establish relations between the nearby utterances that preserve both inter-speaker and intra-speaker dependence on utterances in a dialogue?

\noindent\textbf{Global Information}: We want to have a unified model that can capture the underlying context and handle its effect on each utterance present in the dialogue. A transformer encoder \cite{vaswani2017attention} architecture is a suitable choice for this goal. Instead of following the conventional sequential encoding by adding positional encodings to the input, in our approach, a simple transformer encoder without any positional encodings leverages the entire context to generate distributed representations (features) efficiently corresponding to each utterance. The transformer facilitates the flow of information from all utterances when predicting emotion for a particular utterance.

\noindent\textbf{Local Information:} The emotion expressed in an utterance is often triggered by the information in neighboring utterances. We establish relations between the nearby utterances in a way that is capable of capturing both inter-speaker and intra-speaker effects of stimulus over the emotion state of an utterance. Our approach comes close to DialogueGCN \cite{ghosal2019dialoguegcn}, and we define a graph where each utterance is a node, and directed edges represent various relations. We define relations (directed edges) between nodes $\mathcal{R}_{ij} = u_i\rightarrow u_j$, where the direction of the arrow represents the spoken order of utterances. We categorize the directed relations into two types, for self-dependent relations between the utterances spoken by the same speaker $\mathcal{R}_{intra}$, and interrelations between the utterances spoken by different speakers $\mathcal{R}_{inter}$. We propose to use Relational GCN \cite{RGCN} followed by a GraphTransformer \cite{GraphTransformer} to capture dependency defined by the relations.


\begin{figure*}[t]
\centering
  \includegraphics[scale=0.37]{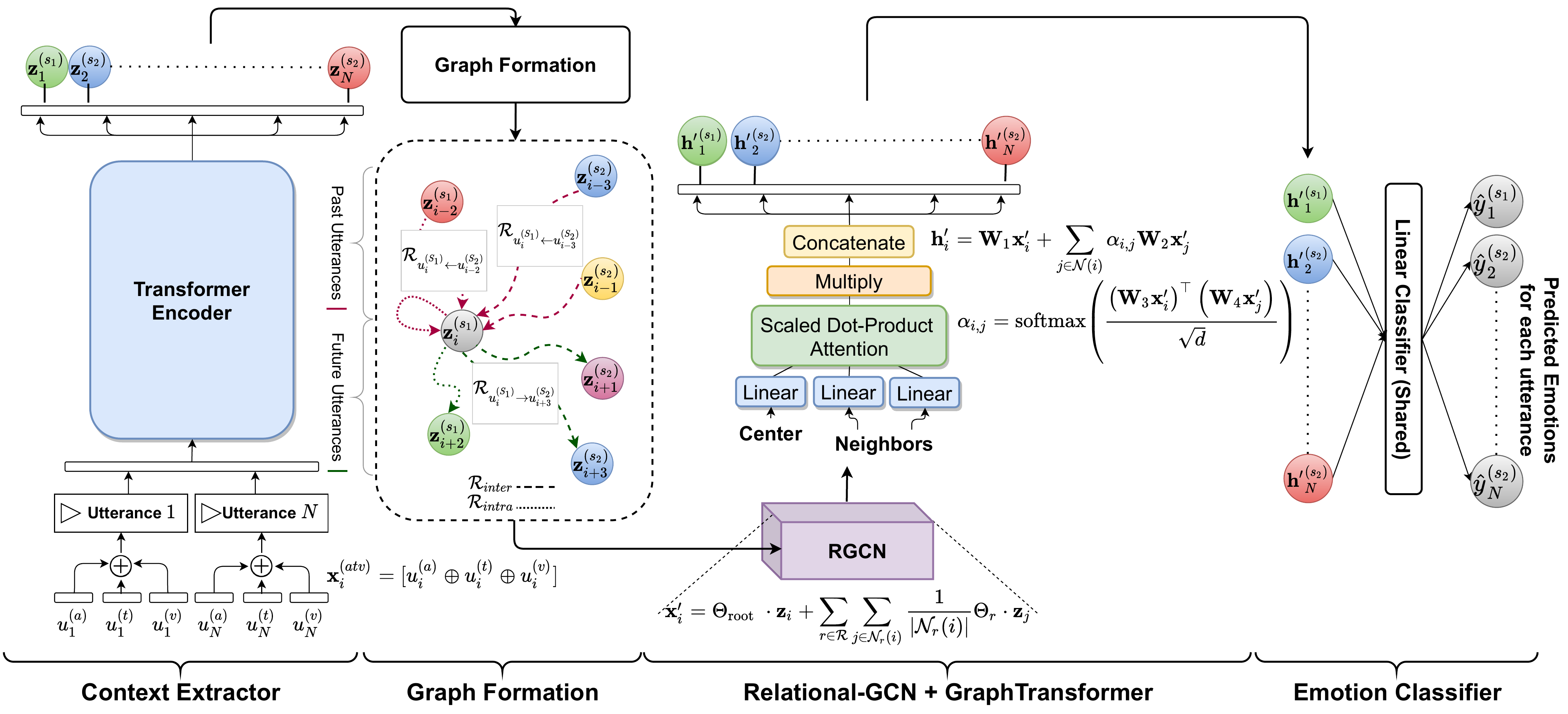}
  \caption{The proposed model (\modelname) architecture. }
  \label{fig:model}
\end{figure*}

\subsection{Overall Architecture}
Figure \ref{fig:model} shows the detailed architecture. The input utterances go as input to the \textbf{Context Extractor} module, which is responsible for capturing the global context. The features extracted for each utterance by the context extractor form a graph based on interactions between the speakers. The graph goes as input to a  \textbf{Relational GCN}, followed by \textbf{GraphTransformer}, which uses the formed graph to capture the inter and intra-relations between the utterances. Finally, two linear layers acting as an \textbf{emotion classifier} use the features obtained for all the utterances to predict the corresponding emotions.

\noindent\textbf{Context Extractor:} Context Extractor takes concatenated features of multiple modalities (audio, video, text) as input for each dialogue utterance ($u_{i}; i = 1,\ldots,n$) and captures the context using a transformer encoder. The feature vector for an utterance $u_{i}$ with the input features corresponding to available  modalities, audio ($u_{i}^{(a)} \in \mathbb{R}^{d_a}$), text ($u_{i}^{(t)} \in \mathbb{R}^{d_t}$) and video ($u_{i}^{(v)} \in \mathbb{R}^{d_v}$) is:  
$$\mathbf{x}_{i}^{(atv)} = [u_{i}^{(a)} \oplus u_{i}^{(t)} \oplus u_{i}^{(v)}] \in  \mathbb{R}^{d}$$ 
where $d=d_a+d_t+d_v$. The combined features matrix for all utterances in a dialogue is given by: 
$$\mathbf{X} = \mathbf{x}^{(atv)} = [\mathbf{x}_{1}^{(atv)},  \mathbf{x}_{2}^{(atv)}  \ldots  ,\mathbf{x}_{n}^{(atv)}]^{T}$$

We define a Query, a Key, and a Value vector for encoding the input features $\mathbf{X} \in \mathbb{R}^{n \times d}$ as follows:
\begin{align*}
Q^{(h)} = \mathbf{X} W_{h, q} , \\
K^{(h)} =  \mathbf{X} W_{h, k}, \\
V^{(h)} =  \mathbf{X} W_{h, v},
\end{align*}
where, $W_{h, q}, W_{h, k}, W_{h, v} \in \mathbb{R}^{d \times k}$

The attention mechanism captures the interaction between the Key and Query vectors to output an attention map $\alpha^{(h)}$, where $\sigma_j$ denotes the softmax function over the row vectors indexed by $j$:
\begin{align*}
\mathbf{\alpha}^{(h)} &= \sigma_j \left(\frac{Q^{(h)}(K^{(h)})^{T}} {\sqrt{k}}\right)
\end{align*}
where $\alpha^{(h)} \in \mathbb{R}^{n \times n}$ represents the attention weights for a single attention head $(h)$.
The obtained attention map is used to compute a weighted sum of the values for each utterance:
\begin{align*}
\text{head}^{(h)} &= \alpha^{(h)} (V^{(h)}) \in \mathbb{R}^{n \times k}\\
\textbf{U}^{'} &= [\text{head}^{(1)} \oplus \text{head}^{(2)} \oplus \ldots \text{head}^{(H)}] W^{o}
\end{align*}
where, $W^{o} \in \mathbb{R}^{kH \times d}$ and $H$ represents the total number of heads in multi-head attention. Note $\textbf{U}^{'} \in \mathbb{R}^{n \times d}$. 
We add residual connection $\mathbf{X}$ and apply $\operatorname{LayerNorm}$, followed by a feed forward and $\operatorname{Add\ \&\ Norm}$ layer:
\begin{align*}
\centering
\mathbf{U} &= \operatorname{LayerNorm}\left(\mathbf{X}+\mathbf{U}^{\prime} ; \gamma_{1}, \beta_{1}\right); \\
\mathbf{Z}^{\prime} &=  \operatorname{ReLU}\left( \mathbf{U} W_{1}\right) W_{2}; \\
\mathbf{Z} &= \operatorname{LayerNorm}\left(\mathbf{U}+\mathbf{Z}^{\prime} ; \gamma_{2}, \beta_{2}\right); 
\end{align*}

where, $\gamma_{1}, \beta_{1} \in \mathbb{R}^{d}$, $W_{1} \in \mathbb{R}^{d \times m}, W_{2} \in \mathbb{R}^{m \times d}$, and $\gamma_{2}, \beta_{2} \in \mathbb{R}^{d}.$
The transformer encoder provides features corresponding to every utterance in a dialogue ($[\mathbf{z}_1, \mathbf{z}_2, \ldots  , \mathbf{z}_n]^T = \mathbf{Z} \in \mathbb{R}^{n \times d}$).\\
\noindent\textbf{Graph Formation:} A graph captures inter and intra-speaker dependency between utterances. Every utterance acts as a node of a graph that is connected using directed relations (past and future relations). We define relation types as speaker to speaker. Formally, consider a conversation between $M$ speakers defined as a dialogue $\mathcal{D}=\{{\mathcal{U}^{S_1}, \mathcal{U}^{S_2},\ldots, \mathcal{U}^{S_M}}\}$, where $\mathcal{U}^{S_1} = \{u^{(S_1)}_1, u^{(S_1)}_2, \ldots, u^{(S_1)}_n\}$ represent the set of utterances spoken by speaker-1. 
We define intra relations between the utterances spoken by the same speaker, $\mathcal{R}_{intra} \in \{\mathcal{U}^{S_i} \rightarrow \mathcal{U}^{S_i}\}$, and inter relations between the utterances spoken by different speakers, $\mathcal{R}_{inter} \in \{ \mathcal{U}^{S_i} \rightarrow \mathcal{U}^{S_j}\}_{i \ne j}$. We further consider a window size and use $\mathcal{P}$ and $\mathcal{F}$ as hyperparameters to form relations between the past $\mathcal{P}$  utterances and future $\mathcal{F}$ utterances for every utterance in a dialogue.
For instance, $\mathcal{R}_{intra}$ and $\mathcal{R}_{inter}$ for utterance $u_i^{(S_1)}$ (spoken by speaker-1) are defined as:

{\small
\begin{align*}\label{eq:intra_inter}
{\mathcal{R}_{intra}(u^{(S_{1})}_{i})} = \{\: u^{(S_1)}_i \leftarrow u^{(S_1)}_{i-\mathcal{P}} \ldots  u^{(S_1)}_i \leftarrow u^{(S_1)}_{i-1}, \\u^{(S_1)}_i \leftarrow u^{(S_1)}_{i}, u^{(S_1)}_i\rightarrow u^{(S_1)}_{i+1} \ldots u^{(S_1)}_{i}\rightarrow u^{(S_1)}_{i+\mathcal{F}} \:\}\\
{\mathcal{R}_{inter} (u^{(S_{1})}_{i})} = \{\: u^{(S_1)}_i \leftarrow u^{(S_2)}_{i-\mathcal{P}},\ldots, u^{(S_1)}_i \leftarrow u^{(S_2)}_{i-1}, \\ u^{(S_1)}_i\rightarrow u^{(S_2)}_{i+1}, \ldots, u^{(S_1)}_{i}\rightarrow u^{(S_2)}_{i+\mathcal{F}} \:\}
\end{align*}
}%

\noindent where $\leftarrow$ and $\rightarrow$ represent the past and future relation type respectively (example in Appendix \ref{app:graph-Formation}).

\noindent\textbf{Relational Graph Convolutional Network (RGCN):} The vanilla RGCN \cite{RGCN} helps accumulate relation-specific transformations of neighboring nodes depending on the type and direction of edges present in the graph through a normalized sum. In our case, it captures the inter-speaker and intra-speaker dependency on the connected utterances.
$$\mathbf{x}_{i}^{\prime}=\Theta_{\text {root }} \cdot \mathbf{z}_{i}+\sum_{r \in \mathcal{R}} \sum_{j \in \mathcal{N}_{r}(i)} \frac{1}{\left|\mathcal{N}_{r}(i)\right|} \Theta_{r} \cdot \mathbf{z}_{j}$$

\noindent where $\mathcal{N}_{r}(i)$ denotes the set of neighbor indices of node $i$ under relation $r \in \mathcal{R}$, $\Theta_{root}$ and $\Theta_{r}$ denote the learnable parameters of RGCN, $|\mathcal{N}_{r}(i)|$ is  the normalization constant and $\mathbf{z}_{j}$ is the utterance level feature coming from the transformer. 


\noindent\textbf{GraphTransformer:} For extracting rich representation from the node features, we use a GraphTransformer \cite{GraphTransformer}. GraphTransformer adopts the vanilla multi-head attention into graph learning by taking into account nodes connected via edges. 
Given node features $H = {\mathbf{x}_1^{\prime}, \mathbf{x}_2^{\prime}, \ldots, \mathbf{x}_n^{\prime}}$ obtained from RGCN, 
$$\mathbf{h}_{i}^{\prime}=\mathbf{W}_{1} \mathbf{x}_{i}^{\prime}+\sum_{j \in \mathcal{N}(i)} \alpha_{i, j} \mathbf{W}_{2} \mathbf{x}_{j}^{\prime} $$
where the attention coefficients $\alpha_{i, j}$  are computed via multi-head dot product attention:
$$\alpha_{i, j}=\operatorname{softmax}\left(\frac{\left(\mathbf{W}_{3} \mathbf{x}_{i}^{\prime}\right)^{\top}\left(\mathbf{W}_{4} \mathbf{x}_{j}^{\prime}\right)}{\sqrt{d}}\right) $$


\noindent\textbf{Emotion Classifier:}
A linear layer over the features extracted by GraphTransformer ($\mathbf{h}_{i}^{\prime}$) predicts the emotion corresponding to the utterance.
$$h_i = \operatorname{ReLU}(W_1 \mathbf{h}_{i}^{\prime} + b_1)$$
$$\mathcal{P}_i = \operatorname{softmax}(W_2 h_i + b_2 )$$
$$\hat{y}_i = \arg \max   (\mathcal{P}_{i} )$$
where $\hat{y}_i$ is the emotion label predicted for the utterance $u_{i}$.

%% file: Experiments.tex
\section{Experiments}
\label{sec:experiments}

\begin{table}[t]\small
\renewcommand{\arraystretch}{1.2}
\setlength\tabcolsep{5pt}
\resizebox{\columnwidth}{!}{
\begin{tabular}{|c|c|c|c|}
\hline
\multirow{2}{*}{\textbf{Dataset}} & \multicolumn{3}{c|}{\textbf{Number of dialogues {[}utterances{]}}} \\ \cline{2-4} 
& \multicolumn{1}{c|}{\textbf{train}} & \multicolumn{1}{c|}{\textbf{valid}} & \textbf{test}    \\ \hline
\textit{\textbf{IEMOCAP}} & \multicolumn{2}{c|}{120 {[}5810 (5146+664){]}}          & 31 {[}1623{]} \\ \hline
\textit{\textbf{MOSEI}}           & \multicolumn{1}{c|}{2249 {[}16327{]}} & \multicolumn{1}{c|}{300 {[}1871{]}} & 646 {[}4662{]} \\ \hline
\end{tabular}
}
\caption{Dataset Statistics.}
\label{tab:dataset_stats}
\end{table}
We experiment for the Emotion Recognition task on the two widely used datasets: \textbf{IEMOCAP} \cite{busso2008iemocap} and \textbf{MOSEI} \cite{zadeh2018multimodal}. The dataset statistics are given in Table \ref{tab:dataset_stats}. IEMOCAP is a dyadic multimodal emotion recognition dataset where each utterance in a dialogue is labeled with one of the six emotion categories: anger, excited, sadness, happiness, frustrated, and neutral. In literature, two IEMOCAP settings are used for testing, one with 4 emotions (anger, sadness, happiness, neutral) and one with 6 emotions. We experiment with both of these settings. MOSEI is a multimodal emotion recognition dataset annotated with 7 sentiments (-3 (highly negative) to +3 (highly positive)) and 6 emotion labels (happiness, sadness, disgust, fear, surprise, and anger). Note that the emotion labels differ across the datasets. We use weighted F1-score and Accuracy as evaluation metrics (details in Appendix \ref{app:metrics}). 

\begin{table*}[t]\small
\centering
\renewcommand{\arraystretch}{1.2}
\setlength\tabcolsep{10pt}
\resizebox{2\columnwidth}{!}{
\begin{tabular}{|c|c|c|c|c|c|c|c|c|}
\hline \multirow{2}{*}{ \textbf{Models} } & \multicolumn{8}{|c|}{ \textbf{IEMOCAP: Emotion Categories} } \\
\cline { 2 - 9 } & \multicolumn{1}{|c|}{ Happy } & \multicolumn{1}{|c|}{ Sad } & \multicolumn{1}{|c|}{ Neutral } & \multicolumn{1}{|c|}{ Angry } & \multicolumn{1}{|c|}{ Excited } & \multicolumn{1}{|c|}{ Frustrated } & \multicolumn{2}{|c|}{ Avg. } \\
\cline { 2 - 9 } &  F1 (\%) &  F1 (\%) &  F1 (\%) &  F1 (\%) &  F1 (\%) &  F1 (\%) & Acc. (\%) & F1 (\%) \\ \hline
\hline bc-LSTM &  $35.6$ &  $69.2$ &  $53.5$ &  $66.3$ &  $61.1$ &  $62.4$ & $59.8$ & $59.0$ \\
\hline memnet &  $33.0$ &  $69.3$ & $55.0$ &  $66.1$ &  $62.3$ &  $63.0$ & $59.9$ & $59.5$ \\
\hline TFN &  $33.7$ &  $68.6$ &  $55.1$ &  $64.2$ &  $62.4$ &  $61.2$ & $58.8$ & $58.5$ \\
\hline MFN &  $34.1$ & $70.5$ &  $52.1$ &  $66.8$ &  $62.1$ &  $62.5$ & $60.1$ & $59.9$ \\
\hline CMN &  $32.6$ &  $72.9$ &  $56.2$ &  $64.6$ &  $67.9$ &  $63.1$ & $61.9$ & $61.4$ \\
\hline ICON &  $32.8$ &  $74.4$ &  $60.6$ &  $6 8 . 2$ &  $68.4$ &  $\textbf{66.2}$ & $64.0$ & $63.5$ \\
\hline DialogueRNN &  $32.8$ &  $7 8 . 0$ &  $59.1$ &  $63.3$ &  $7 3 . 6$ & $59.4$ & $63.3$ & $62.8$ \\
\hline CAN &  $31.8$ & $71.9$ &  $60.4$ &  $66.7$ &  $68.5$ &  $66.1$ & $63.2$ & $62.4$ \\
\hline Af-CAN &  $3 7 . 0$ &  $72.1$ &  $6 0 . 7$ &  $\textbf{67.3}$ & $66.5$ &  $66.1$ &  $6 4 . 6$ & $6 3 . 7$ \\
\specialrule{1pt}{1pt}{1pt}
\modelname &  $\textbf{51.9}$ &  $\textbf{81.7}$ &  $\textbf{68.6}$ & $66.0$ & $\textbf{75.3}$ &  $58.2$ & $\textbf{68.2}$ & $\textbf{67.6}$ \\
\hline
\end{tabular}
}
\caption{Results on IEMOCAP (6-way) multimodal (A+T+V) setting. Avg. denotes weighted average.}
\label{tab:iemo6_all}
\end{table*}

\noindent\textbf{Implementation Details:} For IEMOCAP, audio features (size 100) are extracted using OpenSmile \cite{10.1145/1873951.1874246_opensmile}, video features (size 512) are taken from \citet{8373812_openface2.0}, and text features (size 768) are extracted using sBERT \cite{reimers2019sentencebert_sbert}. 
Audio features for the MOSEI dataset were taken from \citet{delbrouck2020transformer_tbje}, which are extracted using librosa \cite{McFee2015librosaAA} with 80 filter banks, making the feature vector size of 80. The video features (size 35) are taken from \citet{zadeh2018multimodal}. The textual features (size 768) are obtained using sBERT. The textual features are sentence-level static features. For Audio and Visual modalities, we use sentence/utterance level features by averaging all the token level features. We fuse the features of all the available modalities (A(audio)+T(text)+V(video): ATV) via concatenation. We also explored other fusion mechanisms (Appendix \ref{app:fusion}). However, concatenation gave the best performance. We conduct a hyper-parameter search for our proposed model using Bayesian optimization techniques (details in Appendix \ref{app:hyperparams}).


\noindent\textbf{Baselines:} We do a comprehensive evaluation of \modelname\  by comparing it with a number of baseline models.  For IEMOCAP, we compare our model with the existing multimodal frameworks (Table \ref{tab:iemo6_all}), which includes DialogueRNN \cite{majumder2019dialoguernn}, bc-LSTM \cite{poria2017context_bclstm}, CHFusion \cite{majumder2018multimodal_chfusion}, memnet \cite{10.5555/2969442.2969512_memnet}, TFN \cite{zadeh2017tensor_TFN}, MFN \cite{zadeh2018memory_MFN}, CMN \cite{hazarika-etal-2018-conversational_CMN}, ICON \cite{hazarika-etal-2018-icon}, and Af-CAN \cite{wang2021contextual_can}. For MOSEI, \modelname\ is compared (Table \ref{tab:mosei_results}) with multimodal models, including Multilogue-Net \cite{shenoy2020multilogue} and TBJE \cite{delbrouck2020transformer_tbje} (details and analysis of baselines in \S \ref{sec:discussion}).

\begin{table}[!h]\small
\renewcommand{\arraystretch}{1.3}
\setlength\tabcolsep{47pt}
\resizebox{\columnwidth}{!}{
\begin{tabular}{|c|c|c|}
\hline 
\textbf{Model}  & \textbf{F1-score (\%)} \\ \hline \hline
bc-LSTM           & 75.13 \\ \hline
CHFusion       & 76.80   \\ \specialrule{1pt}{1pt}{1pt}
\modelname  & \textbf{84.50}\\ \hline
\end{tabular}
}
\caption{Results on IEMOCAP dataset for 4 emotion classes in multimodal setting (weighted F1-score).}
\label{tab:iemo4_results}
\end{table}

%% file: Results-Analysis.tex
\section{Results and Analysis}{\label{sec:result_analysis}}


\noindent\textbf{IEMOCAP:} Table \ref{tab:iemo6_all} shows the results for IEMOCAP (6-way) multimodal setting. Overall, \modelname\ performs better than all the previous baselines as measured using accuracy and F1-score. We also see an improvement in the class-wise F1 for happy, sad, neutral, and excited emotions. This improvement is possibly due to the GNN architecture (described in analysis later) that we are using in our model, and none of the previous multimodal baselines uses GNN in their architecture. Results for IEMOCAP (4-way) setting are in Table \ref{tab:iemo4_results}. In this setting, \modelname\ achieves 7.7\% improvement over the previous SOTA model.


\begin{table*}[!h]
\renewcommand{\arraystretch}{1.5}
\setlength\tabcolsep{2pt}
\hspace{0.2cm}
\resizebox{2\columnwidth}{!}{
\begin{tabular}{cc|c|c|c|c|c|c|c|c|c|c|c|c|c|c|}
\cline{3-16}
\multicolumn{2}{c|}{} &
  \multicolumn{2}{c|}{\textbf{\begin{tabular}[c]{@{}c@{}}Sentiment Class\\ Accuracy(\%)\end{tabular}}} &
  \multicolumn{6}{c|}{\textbf{\begin{tabular}[c]{@{}c@{}}Emotion Class\\ (weighted) F1-score (\%)\end{tabular}}} &
  \multicolumn{6}{c|}{\textbf{\begin{tabular}[c]{@{}c@{}}Multi-label Emotion Class\\ (weighted) F1-score (\%)\end{tabular}}} \\ \hline \hline
\multicolumn{2}{|c|}{\textbf{Model}} &
  \textbf{2 Class} &
  \textbf{7 Class} &
  \textbf{Happiness} &
  \textbf{Sadness} &
  \textbf{Angry} &
  \textbf{Fear} &
  \textbf{Disgust} &
  \textbf{Surprise} &
  \textbf{Happiness} &
  \textbf{Sadness} &
  \textbf{Angry} &
  \textbf{Fear} &
  \textbf{Disgust} &
  \textbf{Surprise} \\ \hline
\multicolumn{1}{|c|}{\textit{\textbf{Multilogue-Net}}} &
  T + A + V &
  82.88 &
  \textbf{44.83} &
  67.84 &
  65.34 &
  67.03 &
  87.79 &
  74.91 &
  \textbf{86.05} &
  70.6 &
  70.7 &
  74.4 &
  86.0 &
  83.4 &
  87.8 \\ \hline
\multicolumn{1}{|c|}{\multirow{3}{*}{\textit{\textbf{TBJE}}}} &
  T &
  81.9 &
  44.2 &
  - &
  - &
  - &
  - &
  - &
  - &
  63.4 &
  65.8 &
  75.3 &
  84.0 &
  84.5 &
  81.4 
   \\ \cline{2-16} 
    \multicolumn{1}{|c|}{} &
  A + T &
    82.4 &
    43.91 &
    65.91 &
    70.78 &
    70.86 &
    87.79 &
    82.57 &
    86.04 &
    65.5 &
    67.9 &
    76.0 &
    \textbf{87.2} &
    84.5 &
    86.1
   \\ \cline{2-16} 

\multicolumn{1}{|c|}{} &
  T + A + V &
  81.5 &
  44.4 &
  - &
  - &
  - &
  - &
  - &
  - &
  64.0 &
  67.9 &
  74.7 &
  84.0 &
  83.6 &
  86.1
   \\ \specialrule{1.5pt}{1pt}{1pt}
  \multicolumn{1}{|c|}{\multirow{3}{*}{\textit{\textbf{\modelname}}}} &
  T &
  84.42 &
  43.50 &
  69.28 &
  70.49 &
  73.04 &
  87.80 &
  83.69 &
  85.83 &
  69.92 &
  72.16 &
  77.34 &
  86.39 &
  \textbf{86.00} &
  88.27 \\ \cline{2-16} 
  \multicolumn{1}{|c|}{} &
  A + T &
  \textbf{85.00} &
  44.31 &
  68.39 &
  \textbf{73.28} &
  74.98 &
  88.08 &
  \textbf{83.90} &
  85.35 &
  69.62 &
  72.67 &
  76.93 &
  86.39 &
  85.35 &
  88.21 \\ \cline{2-16} 
\multicolumn{1}{|c|}{} &
  T + A + V &
  84.34 &
  43.90 &
  \textbf{70.42} &
  72.31 &
  \textbf{76.20} &
  \textbf{88.17} &
  83.69 &
  85.28 &
  \textbf{72.74} &
  \textbf{73.90} &
  \textbf{78.04} &
  86.71 &
  85.48 &
  \textbf{88.37} \\
  \hline
\end{tabular}
}
\caption{Results on MOSEI dataset. For emotion classification, a weighted F1-score is used. For Sentiment Classification, the results are reported using accuracy. 2 class sentiment consists of only positive and negative sentiment. 7 class sentiment consists of sentiments from highly negative (-3) to highly positive (+3). For the cells showing `-', the results were not provided in the paper, and we were not able to reproduce the results since \textit{TBJE} used token level features, and we are using sentence-level features.}
\label{tab:mosei_results}
\end{table*}

\begin{table}[!h]\small
\renewcommand{\arraystretch}{1.4}
\setlength\tabcolsep{20pt}
\resizebox{\columnwidth}{!}{
\centering
\begin{tabular}{|c|c|}
\hline
\textbf{\# Utterances in Context} & \textbf{F1-score (\%)}\\ \hline \hline
All Utterances in a dialogue & 84.50 \\ \hline
10 Utterances in a dialogue & 77.43 ($\downarrow$7.07)\\ \hline
3 Utterances in a dialogue & 75.39 ($\downarrow$9.11)  \\ \hline
\end{tabular}
}
\caption{Importance of Context in a dialogue. Experiment performed on IEMOCAP (4-way).
}
\label{tab:iemo4_context_importance}
\end{table}

\begin{table}[!h]\small
\renewcommand{\arraystretch}{1.3}
\setlength\tabcolsep{6pt}
\resizebox{\columnwidth}{!}{
\begin{tabular}{c|c|c|c|c|}
\cline{2-5}
        & \textbf{Modalities}    & \textbf{T}             & \textbf{A+T}            & \textbf{A+T+V}           \\ \hline \hline
\multicolumn{1}{|c|}{\multirow{3}{*}{\textbf{(6 way)}}} & Actual        & 66.00         & 65.42         & 67.63         \\ \cline{2-5} 
\multicolumn{1}{|c|}{}                         & w/o GNN       & 64.34 (↓1.66) & 61.69 (↓3.73) & 62.96 (↓4.14) \\ \cline{2-5} 
\multicolumn{1}{|c|}{}                         & w/o Relations &  60.49 (↓5.51)             & 65.32 (↓0.10)              & 62.13 (↓5.50)         \\ \specialrule{1pt}{1pt}{1pt}
\multicolumn{1}{|c|}{\multirow{3}{*}{\textbf{(4 way)}}} & Actual        & 81.55         & 81.59         & 84.50         \\ \cline{2-5} 
\multicolumn{1}{|c|}{}                         & w/o GNN       & 81.18 (↓0.37) & 80.16 (↓1.43) & 80.28 (↓4.22) \\ \cline{2-5} 
\multicolumn{1}{|c|}{}                         & w/o Relations & 76.76 (↓4.79)               &  80.27 (↓1.32)             & 79.61 (↓4.88)         \\ \hline
\end{tabular}
}
\caption{Ablation study on IEMOCAP dataset. All values are F1-score (\%). The results shows the importance of GCN layer.
}
\label{tab:iemo_ablation}
\end{table}

\noindent\textbf{MOSEI:} For emotion classification across 6 emotion classes, we used two settings (as done in previous works): \textit{Binary Classification} across each emotion label where a separate model is trained for every emotion class, and \textit{Multi-label Classification} in which the sentence is tagged with more than 1 emotion and single model predicts multiple classes. The reason for doing this was that Multilogue-Net provides results on binary classification setting and TBJE provides results on Multi-label setting. We ran both models on these settings. For a fair comparison, we use the same utterance level textual features similar to our setting (extracted from sBERT) and train Multilogue-Net architecture on both the settings. Originally, Multilogue-Net used GloVe embeddings \cite{pennington-etal-2014-glove} for textual features, and actual results in the paper are different than reported here. For TBJE, we use the features provided by the paper as it uses token-level features. \modelname\ outperforms (Table \ref{tab:mosei_results}) the baseline models in most of the cases. For 2 class sentiment classification, \modelname\ outperforms the previous baselines with the highest accuracy score of 85\% for \textit{A+T}. For 7 class, our model shows comparable performance. All the multimodal approaches tend to perform poorly when adding visual modality, possibly because of noise present in the visual modality and lack of alignment with respect to other modalities. In contrast, our model can capture rich relations across the modalities and show a performance boost while adding visual modality. 

\begin{table}[h]\small
\renewcommand{\arraystretch}{1.4}
\setlength\tabcolsep{25pt}
\resizebox{\columnwidth}{!}{
\begin{tabular}{|c|c|c|}
\hline 
\textbf{Model} & \textbf{Modality} & \textbf{F1-score (\%)}  \\ \hline 
\multicolumn{3}{|c|}{\textbf{\:\:\:\:\:\:\:4-way}} \\ \hline
DialogueGCN & T     & 71.58  \\ \hline
DialogXL    & T     & 73.02  \\ \hline
DAG-ERC     & T     & 78.08   \\ \hline
\multirow{2}{*}{\modelname}  & T     & \textbf{81.55}  \\ 
            & A+T+V & \textbf{84.50} \\ \hline
\multicolumn{3}{|c|}{\textbf{\:\:\:\:\:\:\:6-way}} \\ \hline
EmoBERTa    & T & \textbf{68.57}\\ \hline
DAG-ERC     & T & 68.03 \\ \hline
CESTa       & T & 67.10 \\ \hline
SumAggGIN   & T & 66.61\\ \hline
DialogueCRN & T & 66.20 \\ \hline
DialogXL    & T & 65.94\\ \hline
DialogueGCN & T & 64.18\\ \hline
\multirow{2}{*}{\modelname}  & T  & 66.00\\ 
  & A+T+V & 67.63\\ \hline
\end{tabular}
}
\caption{Comparison with unimodal architectures on IEMOCAP dataset.}
\label{tab:iemo_unimodal}
\end{table}


\noindent We conducted further analysis on our model. Although due to space limitations, the results below mainly describe experiments over IEMOCAP, similar trends were observed for MOSEI as well. 

\noindent\textbf{Effect of Local and Global Info.:} We test our architecture in two information utilization settings: global and local. To test the importance of context in our architecture, we create a sub-dataset using the IEMOCAP (4-way) setting by splitting each dialogue into $n$ utterances and training our architecture. Table \ref{tab:iemo4_context_importance} shows the decrease in performance with number of utterances present in a dialogue (more details on effect of window size in Appendix \ref{app:windowSize}). This experiment helps understand the importance of context in a dialogue. Moreover, it points towards challenges in developing a real-time system (details in \S \ref{sec:discussion}). We test the local information hypothesis by removing the GNN module and directly passing the context extracted features to the emotion classifier. Table \ref{tab:iemo_ablation} shows the drop in performance across modalities when the GNN component is removed from the architecture, making our local information hypothesis more concrete.

\noindent\textbf{Effect of Relation Types:} We also test the effect of inter and intra-relations in the dialogue graph by making all relations of the same type and training the architecture. We observe a drop in performance (Table \ref{tab:iemo_ablation}) when the relations are kept the same in the graph formation step. The explicit relation formation helps capture the local dependencies present in the dialogue. 

\noindent\textbf{Effect of Modalities:} The focus of this work is multimodal emotion recognition. However, just for the purpose of comparison, we also compare with unimodal (text only) approaches. We compare (Table \ref{tab:iemo_unimodal}) with EmoBERTa \cite{kim2021emoberta}, DAG-ERC \cite{shen2021directed_dagerc}, CESTa \cite{wang2020contextualized_cesta}, SumAggGIN \cite{sheng2020summarize_sumaggin}, DialogueCRN \cite{hu2021dialoguecrn}, DialogXL \cite{shen2020dialogxl} and DialogueGCN \cite{ghosal2019dialoguegcn}. Text-based models are specifically optimized for text modalities and incorporate changes to architectures to cater to text. It is not fair to compare with our multimodal approach from that perspective. As shown in results, \modelname, being a fairly generic architecture, still gives better (for IEMOCAP (4-way)) or comparable performance with respect to the SOTA unimodal architectures. In the case of our model, adding more information via other modalities helps to improve the performance. Results on different modality combinations are in Appendix \ref{app:iemocap-diff-modalities}.

\begin{figure}[!h]
\centering
  \includegraphics[scale=0.28]{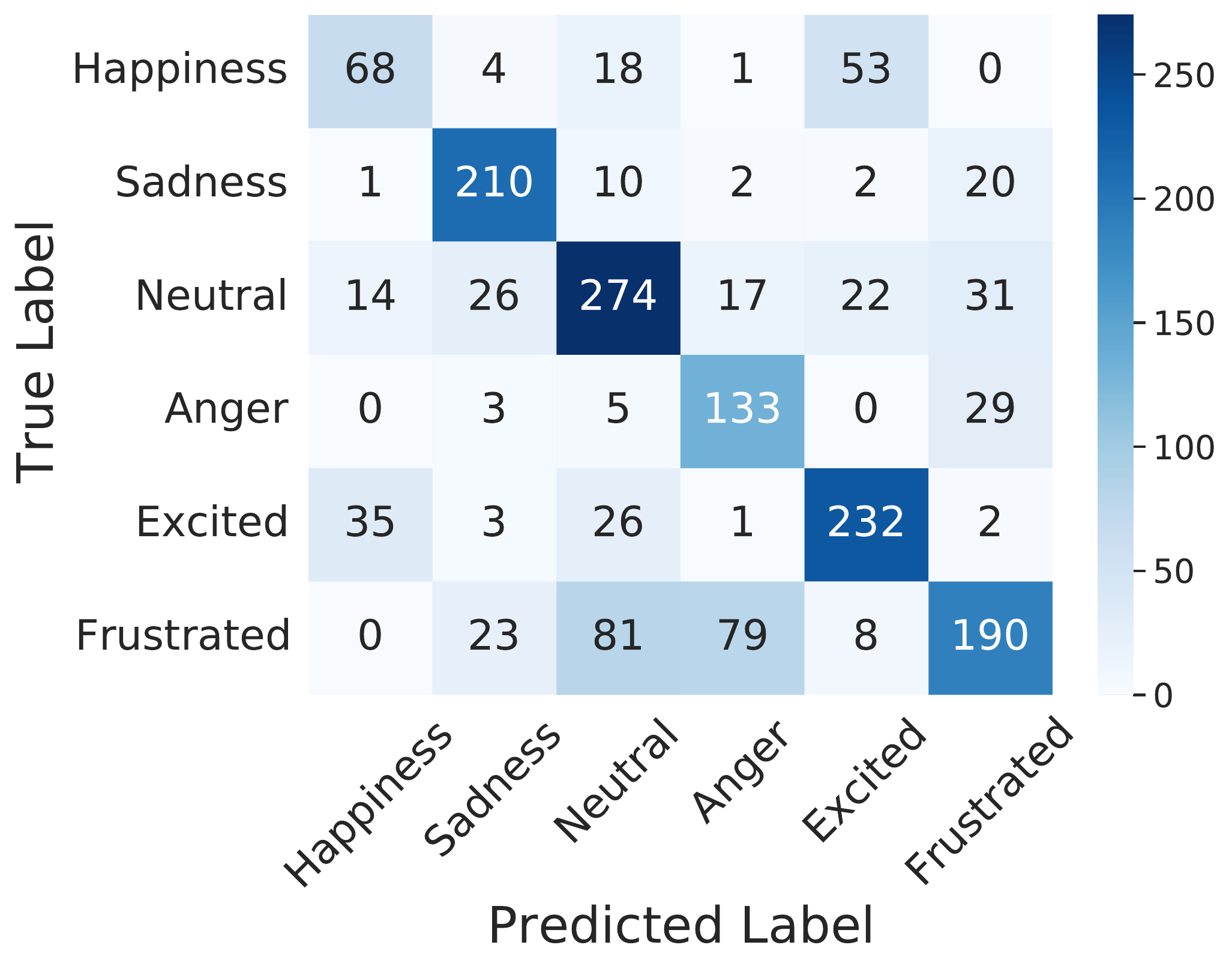}
  \caption{Confusion Matrix for IEMOCAP (6-way)}
  \label{fig:cm6}
\end{figure}

\noindent\textbf{Error Analysis:} After analysing the predictions made across the datasets, we find that our model falls short in distinguishing between similar emotions, such as \textit{happiness} vs \textit{excited} and \textit{anger} vs \textit{frustration} (Figure \ref{fig:cm6}). This issue also exists in previous methods as reported in \citet{shen2021directed_dagerc}, and \citet{ghosal2019dialoguegcn}. We also find that our model misclassifies the other emotion labels as neutral because of a more significant proportion of neutral labeled examples. Moreover, we observe the accuracy of our model when classifying examples having emotion shift is $53.6\%$ compared to $74.2\%$ when the emotion remains the same (more details in Appendix \ref{app:dataset-analysis}). 

\begin{figure}[!h]
\centering
  \includegraphics[scale=0.24]{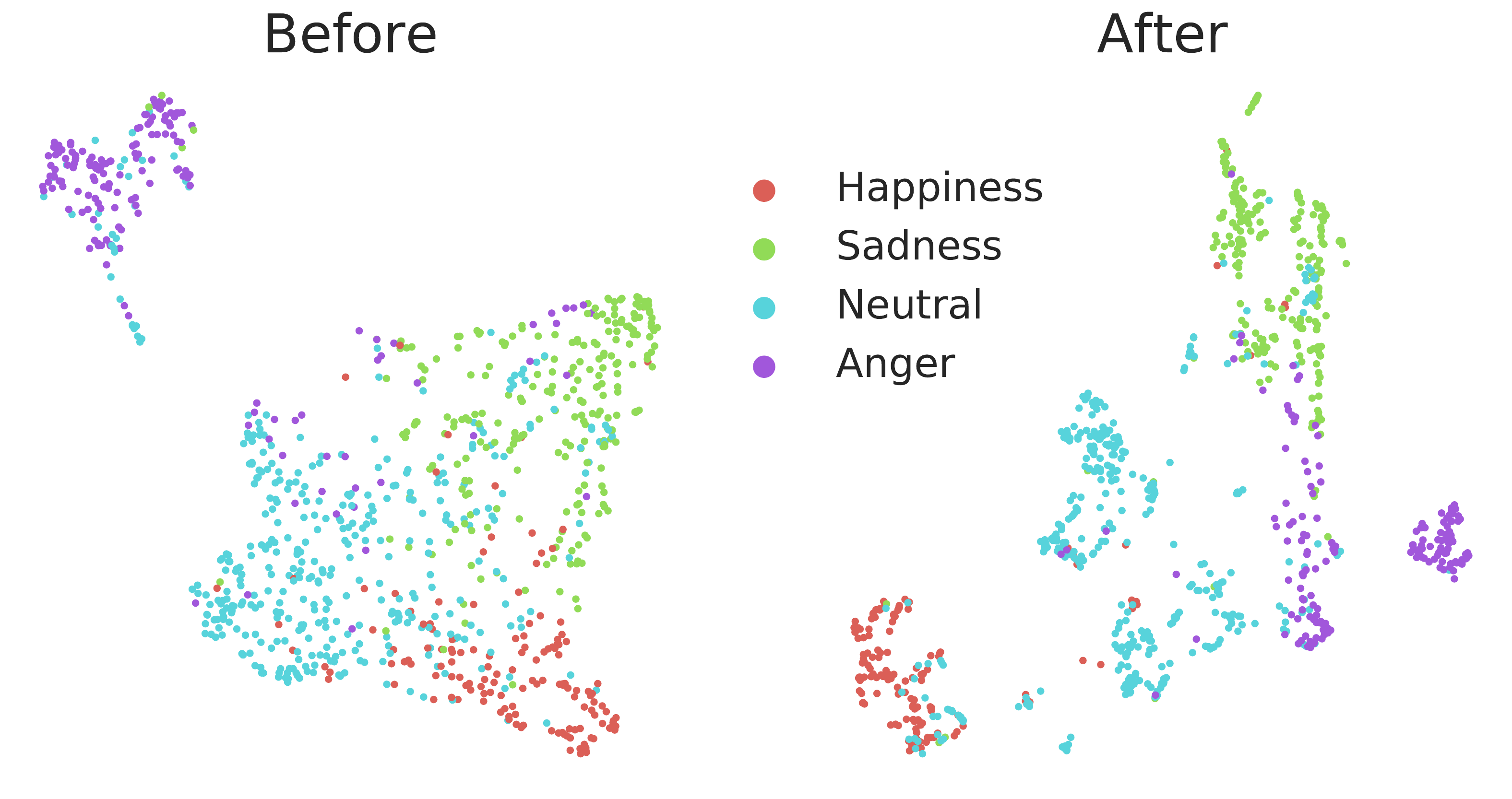}
  \caption{UMAP \cite{becht2019dimensionality_umap}  representation of IEMOCAP (4-way) features before and after GNN.}
  \label{fig:bef_aft_gnn4}
\end{figure}

\noindent\textbf {Efficacy of the GNN Layer:} For observing the effect of the GNN component in our architecture, we also visualize the features before and after the GNN component.  Figure \ref{fig:bef_aft_gnn4} clearly shows the better formation of emotion clusters depicting the importance of capturing local dependency in utterances for better performance in emotion recognition (more in Appendix \ref{app:addi-analysis} and Appendix Figure-\ref{app-fig:bef-aft-gnn6}). 

\begin{figure}[!h]
\centering
  \includegraphics[scale=0.26]{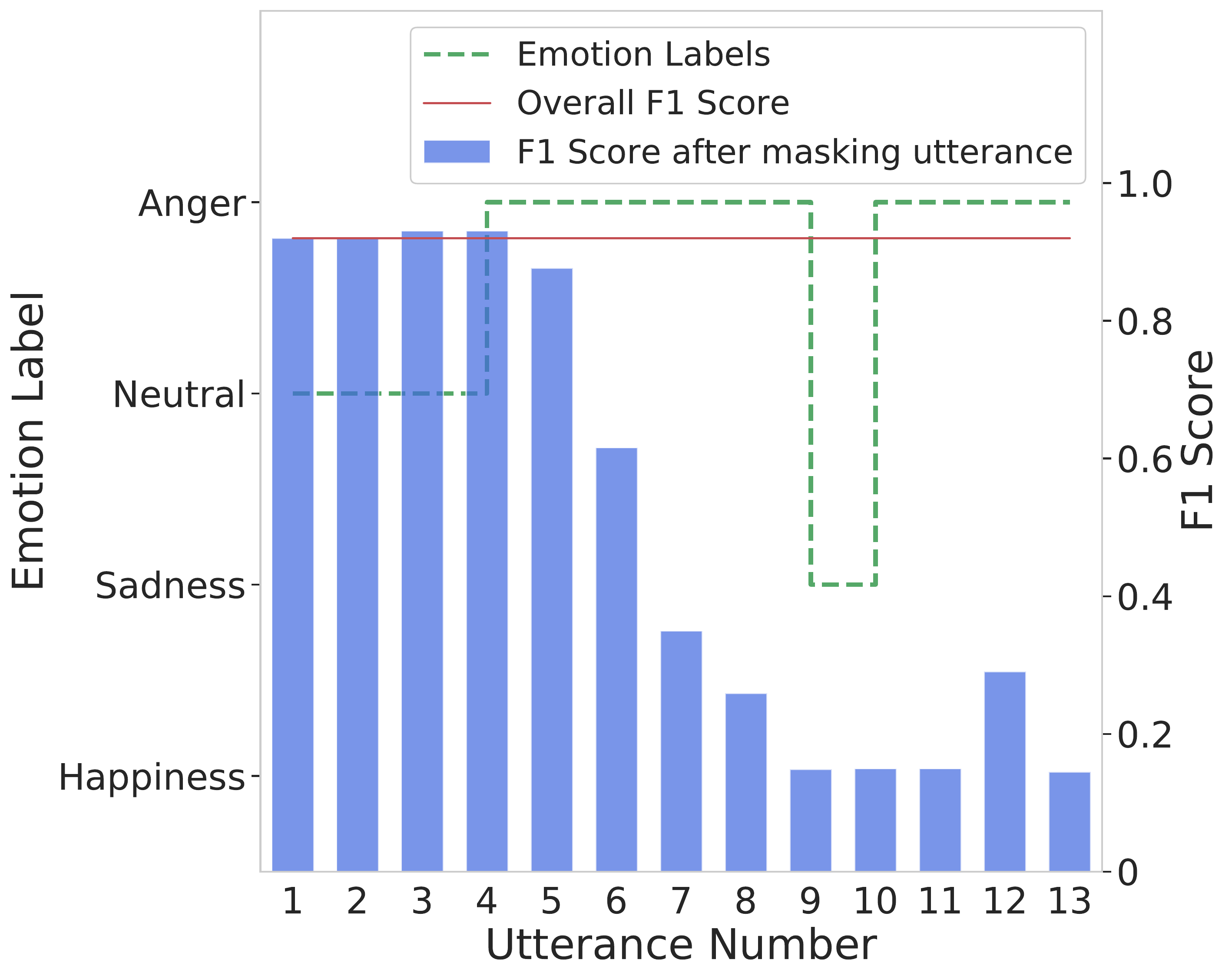}
  \caption{Importance of utterances in IEMOCAP (4-way). Performance drop is observed while masking $9^{th}, 10^{th}$ and $11^{th}$ utterances during inference.}
  \label{fig:importance-of-utterances}
\end{figure}

\noindent\textbf {Importance of utterances:} To verify the effect of utterances and their importance in a prediction for a dialogue, we infer the trained model on dialogues by masking one utterance at a time and calculating the F1-score for prediction. Figure \ref{fig:importance-of-utterances} shows the obtained results for a dialogue (Appendix Table \ref{app:tab:utterances-fig-5}) instance taken randomly from IEMOCAP (4-way) (more in Appendix \ref{app:addi-analysis}). For the first 4 utterances, emotions state being neutral, the effect of masking the utterances is significantly less. In contrast, masking the utterances with emotion shift (9, 10, 11) completely drops the dialogue's F1-score, showing that our architecture captures the effects of emotions present in the utterances.

%% file: Discussion.tex
\section{Discussion} \label{sec:discussion}
\noindent\textbf{Comparison with Baselines:} \label{sec:compBaselines}
Emotion recognition in a multimodal conversation setting comes with two broadly portrayed research challenges \cite{poria2019emotionResearchChallenges}, first, the ability of a model to capture global and local context present in the dialogues, and second, the ability to maintain self and interpersonal dependencies among the speakers. All the popular baselines like Dialogue-GCN \cite{ghosal2019dialoguegcn}, DialogueRNN \cite{majumder2019dialoguernn}, bc-LSTM \cite{poria2017context_bclstm} Af-CAN \cite{afCAN}, etc., try to address these challenges by proposing various architectures. bc-LSTM (bi-directional contextual LSTM \cite{poria2017context_bclstm}) uses LSTM to capture the contextual information and maintain long relations between the utterances from the past and future. Another contemporary architecture Af-CAN \cite{afCAN} utilizes recurrent neural networks based on contextual attention to model the interaction and dependence between speakers and uses bi-directional GRU units to capture the global features from past and future. We propose to address these issues using a unified architecture that captures the effect of context on utterances while maintaining the states for self and interpersonal dependencies. We make use of transformers for encoding the global context and make use of GraphTransformers to capture the self and interpersonal dependencies. Our way of forming relational graphs between the utterances comes close to DialogueGCN (unimodal architecture). We further use a shared Emotion classifier for predicting emotions from all the obtained utterance level features. Moreover, our unified architecture handles multiple modalities effectively and shows an increase in performance after adding information from other modalities.\\
\noindent\textbf{Limitations (Offline Setting):}
A noteworthy limitation of all the proposed Emotion Recognition approaches (including the current one) is that they use global context from past and future utterances to predict emotions. However, baseline systems compared in this paper are also offline systems. For example,  bc-LSTM (bi-directional contextual LSTM) and Af-CAN use utterances from the past and future to predict emotions. Other popular baselines like DialogueGCN and DialogueRNN (BiDialogueRNN) also peek into the future, assuming the presence of all the utterances during inference (offline setting). All such systems that depend on future information can only be used in an offline setting to process and tag the dialogue. An Emotion Recognition system that could work in an online setting exhibits another line of future work worth exploring due to its vast use cases in live telecasting and telecommunication. A possible approach to maintain the context in an online setting would be to take a buffer of smaller context size, where the model can predict emotions taking not the complete dialogue but a smaller subset of it as input in real-time. We tried exploring this setting for our architecture with an online buffer of maintaining a smaller context window. For experimenting with it, we created a sub-dataset using the IEMOCAP (4-way) setting by splitting each dialogue into $n$ utterances and training our architecture. Our results in Table \ref{tab:iemo4_context_importance} show the decrease in performance with the number of utterances present in a dialogue depicting the importance of context in a conversation. Performance improvements in these settings where the system can work in real-time are worth exploring and are an interesting direction for future research.

%% file: Conclusion.tex
\section{Conclusion and Future Work}
\label{sec:conclusion}
We present a novel approach of using GNNs for multimodal emotion recognition and propose \textbf{\modelname: COntextualized GNN based Multimodal Emotion recognitioN}. We test \modelname\ on two widely known multimodal emotion recognition datasets, IEMOCAP and MOSEI. \modelname\ outperforms the existing state-of-the-art methods in multimodal emotion recognition by a significant margin (i.e., 7.7\% F1-score increase for IEMOCAP (4-way)). By comprehensive analysis and ablation studies over \modelname, we show the importance of different modules. \modelname\ fuses information effectively from multiple modalities to improve the performance of emotion prediction tasks. We perform a detailed error analysis and observe that the misclassifications are mainly between the similar classes and emotion shift cases. We plan to address this in future work, where the focus will be to incorporate a component for capturing the emotional shifts for fine-grained emotion prediction. 


%% file: Acknowledgements.tex
\section{Acknowledgements}
We would like to thank reviewers for their insightful comments. This research is supported by SERB India (Science and Engineering Board) Research Grant number SRG/2021/000768. 

%% file: Appendix.tex
\section*{Appendix}
\label{appendix}



\section{Hyperparameter Setting}\label{app:hyperparams}

Hyperparameters used to train our model are described in Table \ref{tab:iemo_hyperparam} for IEMOCAP (4-way and 6-way) and Table \ref{tab:mosei_hyper} for MOSEI dataset.

\begin{table}[!h]
\renewcommand{\arraystretch}{1.5}
\setlength\tabcolsep{20pt}
\resizebox{\columnwidth}{!}{
\begin{tabular}{c|c|c|c}
\hline
\textbf{Dropout} & \textbf{GNNHead} & \textbf{SeqContext} & \textbf{ILR} \\ \hline \hline
0.1 & 7 & 4 & 1e-4 \\ \hline
\end{tabular}
}
\caption{\textbf{Hyperparameter values for our model on IEMOCAP dataset.} \textit{ILR: Initial learning rate.}}
\label{tab:iemo_hyperparam}
\end{table}


\begin{table}[!h]\small
\renewcommand{\arraystretch}{1.5}
\setlength\tabcolsep{12pt}
\resizebox{\columnwidth}{!}{
\centering
\begin{tabular}{c|c|c|c|c}
\hline
\textbf{Modalities} & \textbf{Dropout} & \textbf{GNNHead} & \textbf{SeqContext} & \textbf{ILR}  \\ \hline
T & 0.399 & 3 & 5 & 3.3e-3  \\ \hline
A+T & 0.103 & 1 & 2 & 6.9e-3  \\ \hline
A+T+V & 0.337 & 2 & 1 & 1.1e-3  \\ \hline
\end{tabular}
}
\caption{\textbf{Hyperparameter value on MOSEI dataset}. \textit{ILR: Initial learning rate}.
}
\label{tab:mosei_hyper}
\end{table}

We use PyTorch \cite{NEURIPS2019_9015_pytorch} for training our architecture and PyG (PyTorch Geometric) \cite{DBLP:journals/corr/abs-1903-02428_pytorchgeometric} for the GNN component in our architecture. We use comet \cite{CometML} for logging all our experiments and its Bayesian optimizer for hyperparameter tuning. Our architecture trained on the IEMOCAP dataset has 55,932,052 parameters and takes around 7 minutes to train for 50 epochs on the NVIDIA Tesla K80 GPU. Comparison of the model with baselines in terms of the number of parameters is challenging, as the baselines parameters vary depending on the hyperparameter setting. Moreover, many baselines do not provide information about the number of parameters. 


\section{Dataset Analysis} \label{app:dataset-analysis}
\begin{figure}[!h]
\centering
  \includegraphics[scale=0.33]{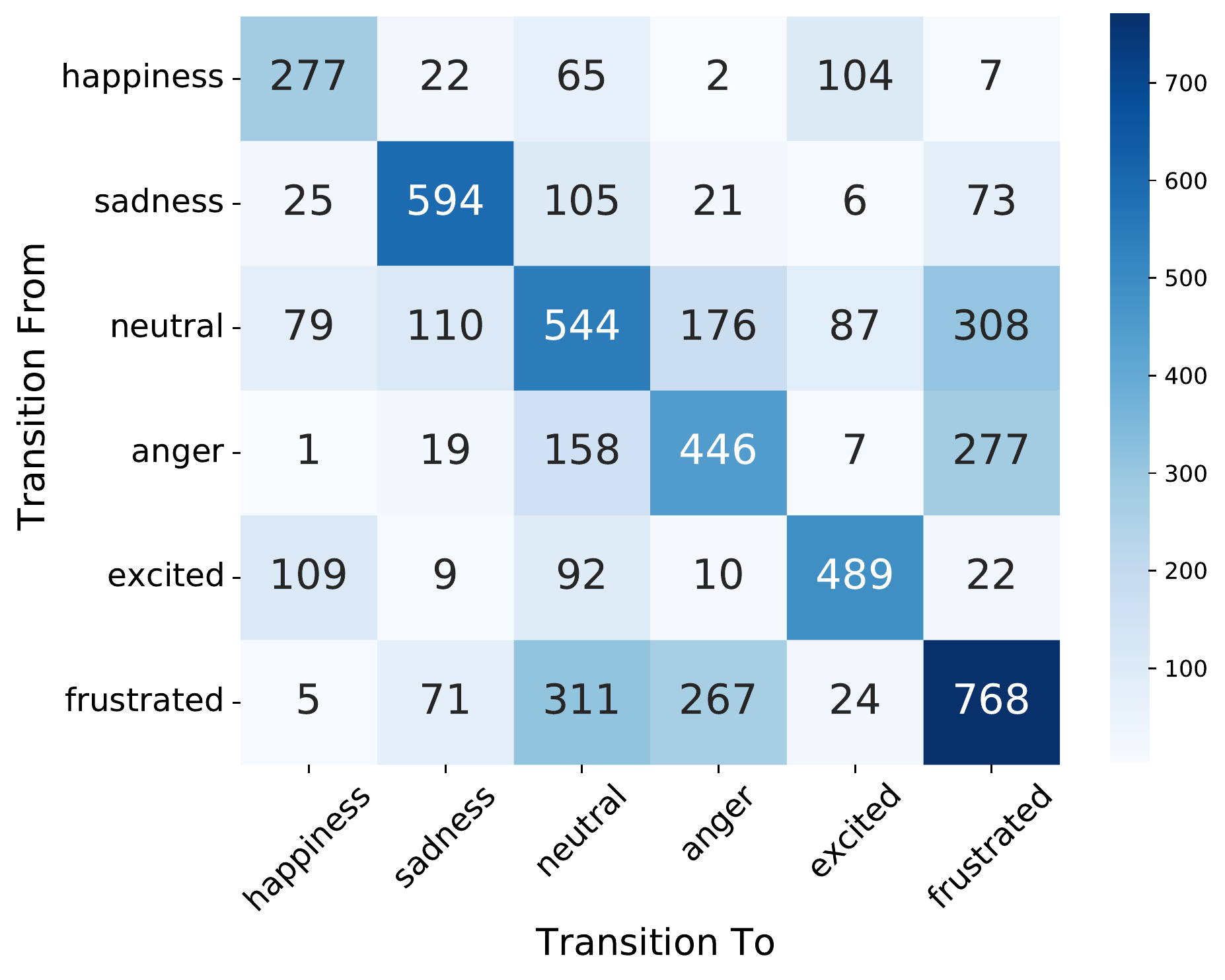}
  \caption{ Utterance-level Emotion transition for IEMOCAP. These are emotions transitions in consecutive utterances across speakers.}
  \label{fig:ut_emo_tran}
\end{figure}

\begin{figure}[!h]
\centering
  \includegraphics[scale=0.33]{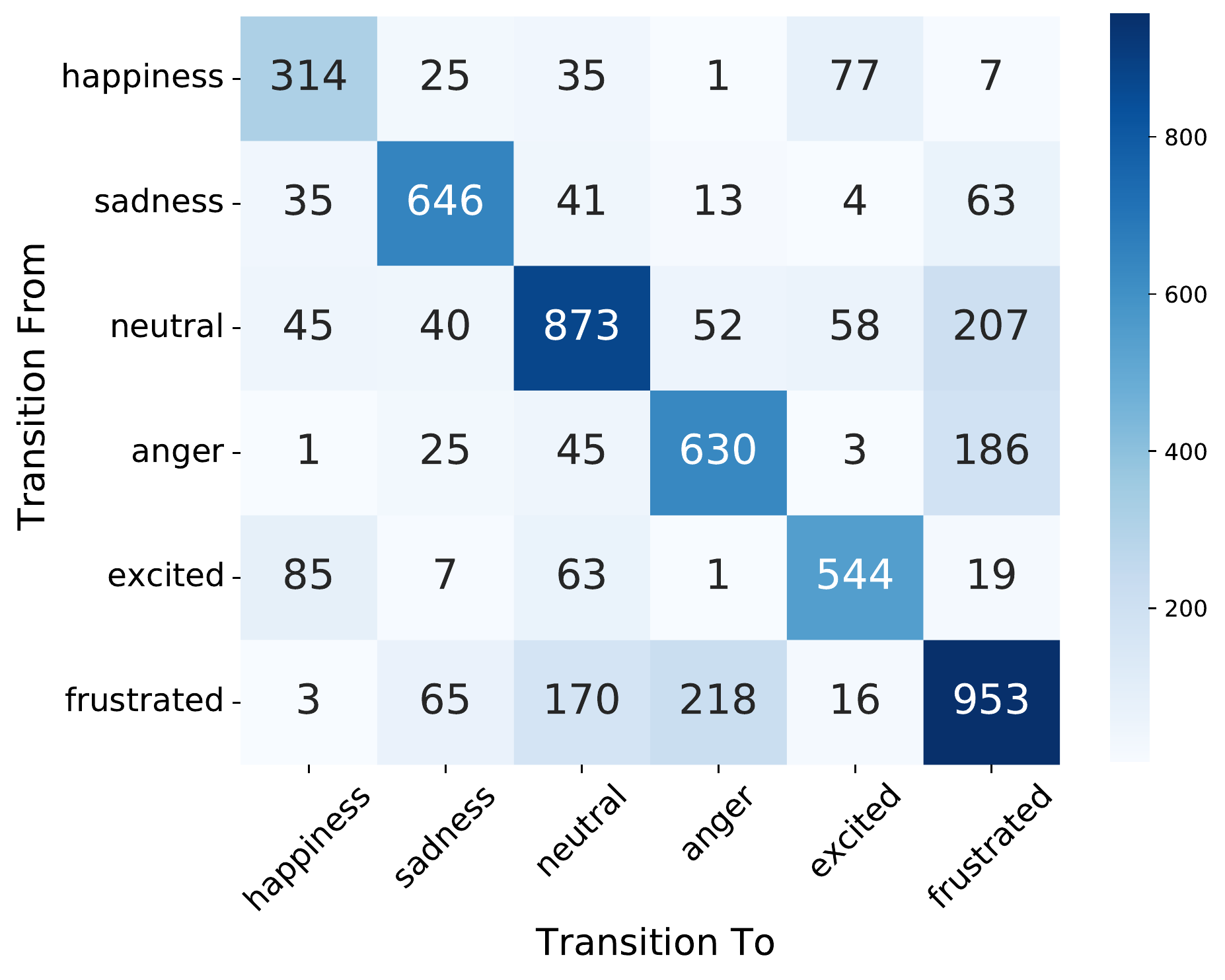}
  \caption{Speaker-level Emotion transition for IEMOCAP. These are emotions transitions in the consecutive utterances of the same speaker.}
  \label{fig:sp_emo_tran}
\end{figure}

We study IEMOCAP dataset in detail for error analysis of our model. We observe the emotion transition at Utterance level (Figure \ref{fig:ut_emo_tran}) and Speaker level (Figure \ref{fig:sp_emo_tran}). We find a high percentage of transitions between similar emotions, causing the models to confuse between the similar classes of emotion. Considering the emotion transition between states that are opposite, like from happy to sad, we deduce the poor performance of emotion recognition architectures for such cases. We plan to address this issue in future work where we target a model which performs better in fine-grained emotion recognition and is robust towards the shifts in emotions.


\section{Evaluation Metrics}
{\label{app:metrics}
\noindent \textbf{Weighted F1 Score:} The F1 score can be interpreted as a harmonic mean of the precision and recall, where an F1 score reaches its best value at 1 and worst score at 0. The relative contribution of precision and recall to the F1 score are equal. The formula for the F1 score is:

$$F1 = 2 * \frac{(precision * recall)}{  (precision + recall)}$$

\noindent For weighted F1 score, we calculate metrics for each label, and find their average weighted by support (the number of true instances for each label). 

\noindent \textbf{Accuracy}: It is defined as the percentage of correct predictions in the test set.
}

\begin{table*}[t]
\centering
\renewcommand{\arraystretch}{1}
\setlength\tabcolsep{20pt}
\resizebox{2\columnwidth}{!}{
\begin{tabular}{|c|l|c|}
\hline
\textbf{Speaker} & \multicolumn{1}{c|}{\textbf{Utterance Text}}                                                                                                                                                                                       & \textbf{Emotion} \\ \hline
M                & 'Why does that bother you?'                                                                                                                                                                                                        & neutral          \\ \hline
F                & "She's been in New York three and a half years. Why all of the sudden?"                                                                                                                                                            & neutral          \\ \hline
M                & 'Well maybe. Maybe she just wanted to see her again.'                                                                                                                                                                              & neutral          \\ \hline
F                & \begin{tabular}[c]{@{}l@{}}"What did you mean? He lived next door to the girl all of his life, \\ why wouldn't he want to see her again? Don't look at me like that, \\ he didn't tell me any more than he told you."\end{tabular} & neutral          \\ \hline
M                & "She's not his girl. She knows she's not."                                                                                                                                                                                         & angry            \\ \hline
F                & "I want you to pretend like he's coming back!"                                                                                                                                                                                     & angry            \\ \hline
M                & "Because if he's not coming back, then I'll kill myself."                                                                                                                                                                          & angry            \\ \hline
F                & \begin{tabular}[c]{@{}l@{}}'Laugh. Laugh at me, but what happens the night that she goes to sleep in his bed, \\ and his memorial breaks in pieces?"\end{tabular}                                                                  & angry            \\ \hline
M                & \begin{tabular}[c]{@{}l@{}}'Only last week, another boy turned up in Detroit, \\ been missing longer than Larry, \\ you read it yourself, '\end{tabular}                                                                           & angry            \\ \hline
F                & "You've got to believe. You've got to--"                                                                                                                                                                                           & sad              \\ \hline
M                & "What do you mean me above all? Look at you. You're shaking!"                                                                                                                                                                      & angry            \\ \hline
F                & "I can't help it!"                                                                                                                                                                                                                 & angry            \\ \hline
M                & 'What have I got to hide? What the hell is the matter with you, Kate?'                                                                                                                                                             & angry            \\ \hline
\end{tabular}
}
\caption{Dialogue utterances corresponding to plot shown in Figure \ref{fig:importance-of-utterances}.} \label{app:tab:utterances-fig-5}
\end{table*}

\section{Results on Modality Combinations} \label{app:iemocap-diff-modalities}

\begin{figure}[!h]
\centering
  \includegraphics[scale=0.46]{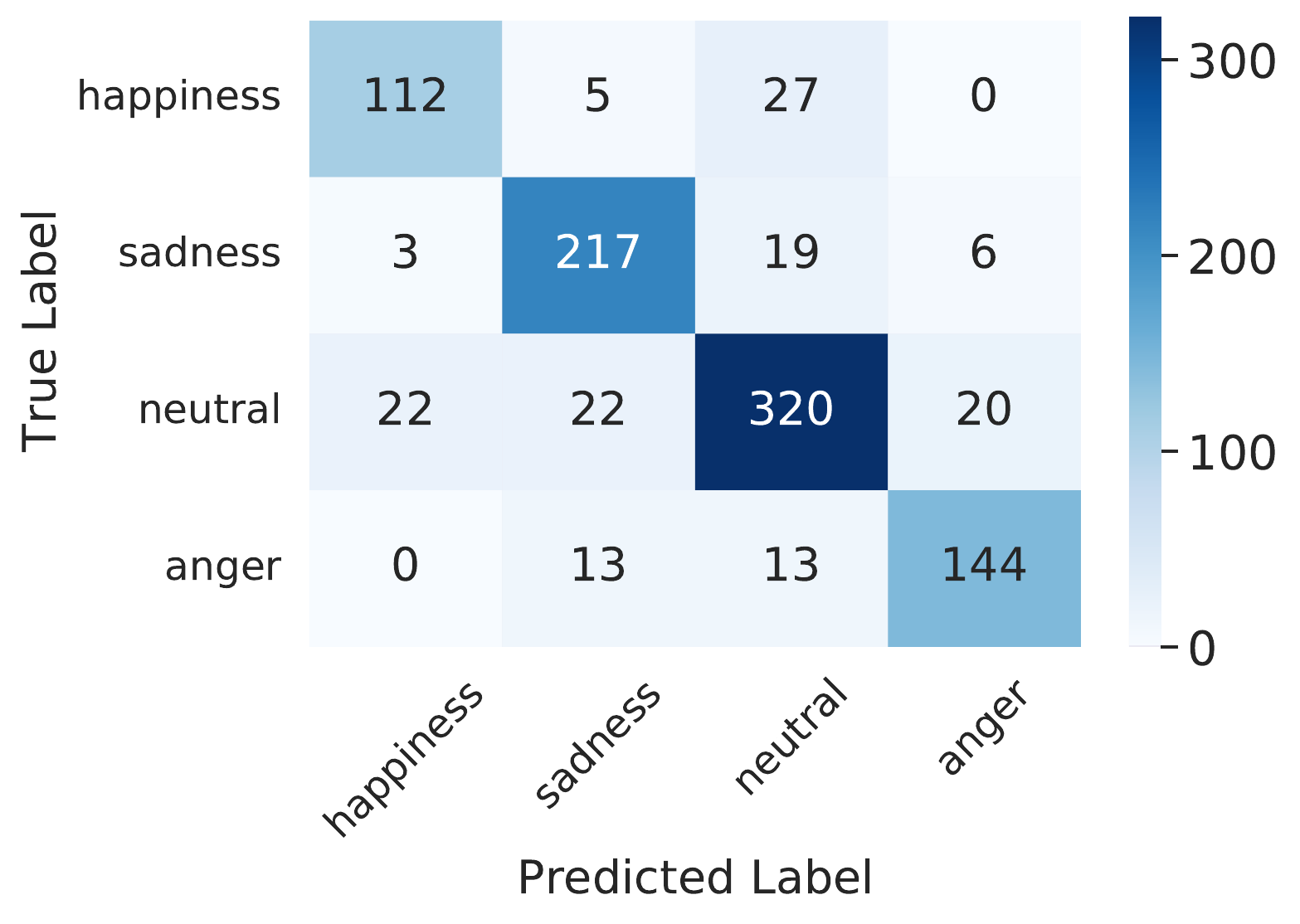}
  \caption{Confusion Matrix for IEMOCAP 4-Way classification }
  \label{fig:cm4}
\end{figure}

Table \ref{tab:iemocap_all_modalities} shows results on the IEMOCAP dataset for all the modality combinations for our architectures. Figure \ref{fig:cm4} shows the confusion matrix for prediction on IEMOCAP 4-way dataset.

\begin{table}[!hbt]\small
\renewcommand{\arraystretch}{1.5}
\setlength\tabcolsep{15pt}
\hspace{0.2cm}
\centering
\resizebox{\columnwidth}{!}{
\begin{tabular}{|c|c|c|}
\hline
\textbf{Modalities}   & \textbf{IEMOCAP-4way} & \textbf{IEMOCAP-6way}\\
                      & F1 Score (\%)         & F1 Score (\%)
\\ \hline
a            & 63.58             &  47.57                   \\ \hline
t            & 81.55             &  66.00                    \\ \hline
v            & 43.85             &  37.58                   \\ \hline
at           & 81.59             &  65.42                    \\ \hline
av           & 64.48             &  52.20                   \\ \hline
tv           & 81.52             &  62.19                   \\ \hline
\textbf{atv} & \textbf{84.50}    &  \textbf{67.63}           \\ \hline
\end{tabular}}
\caption{Results on IEMOCAP-4way and IEMOCAP-6way datasets}
\label{tab:iemocap_all_modalities}
\end{table}

\begin{figure}[!h]
\centering
  \includegraphics[width=\linewidth]{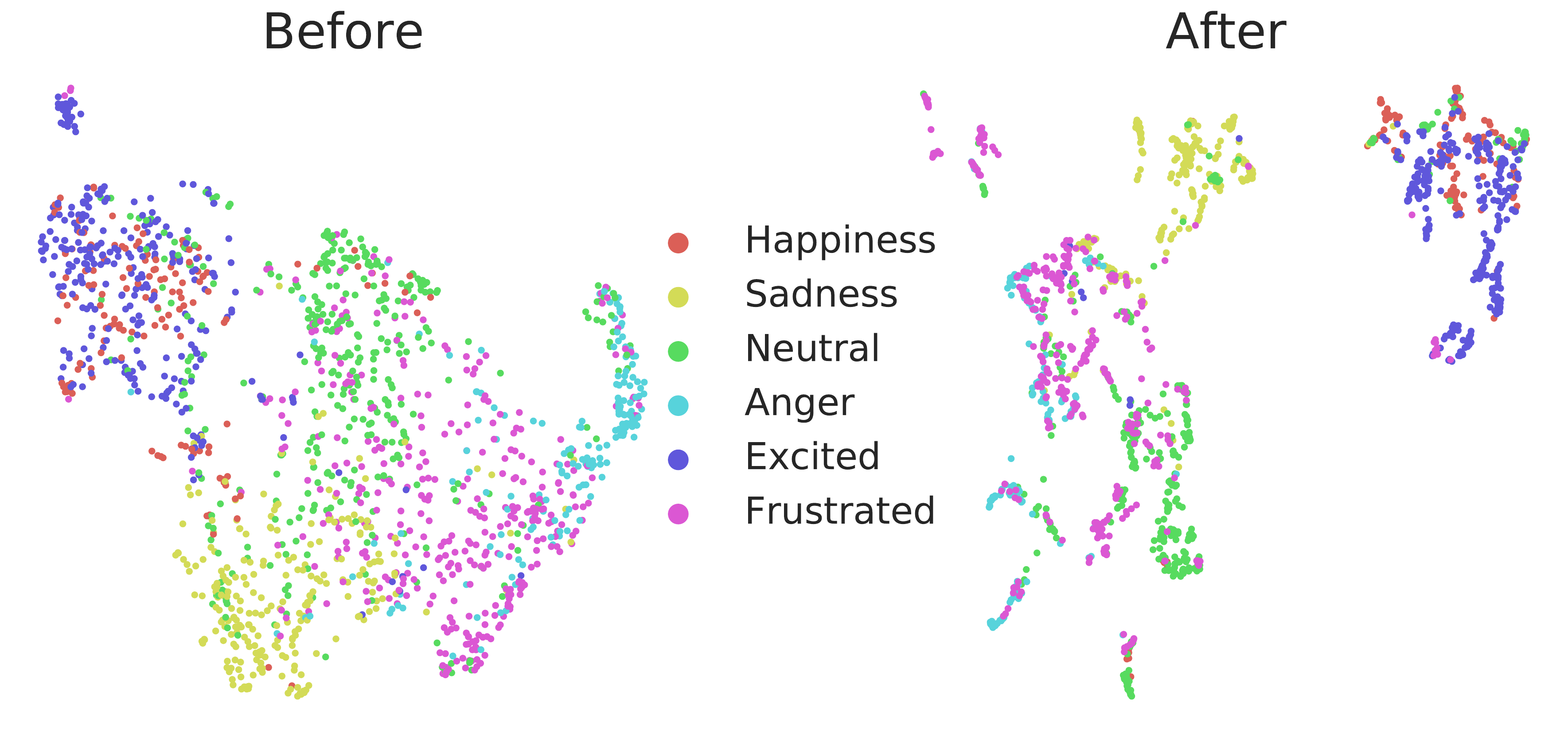}
  \caption{UMAP \cite{becht2019dimensionality_umap}  representation of IEMOCAP 6-way features before and after GNN.}
  \label{app-fig:bef-aft-gnn6}
\end{figure}

\begin{figure}[!h]
\centering
  \includegraphics[scale=0.27]{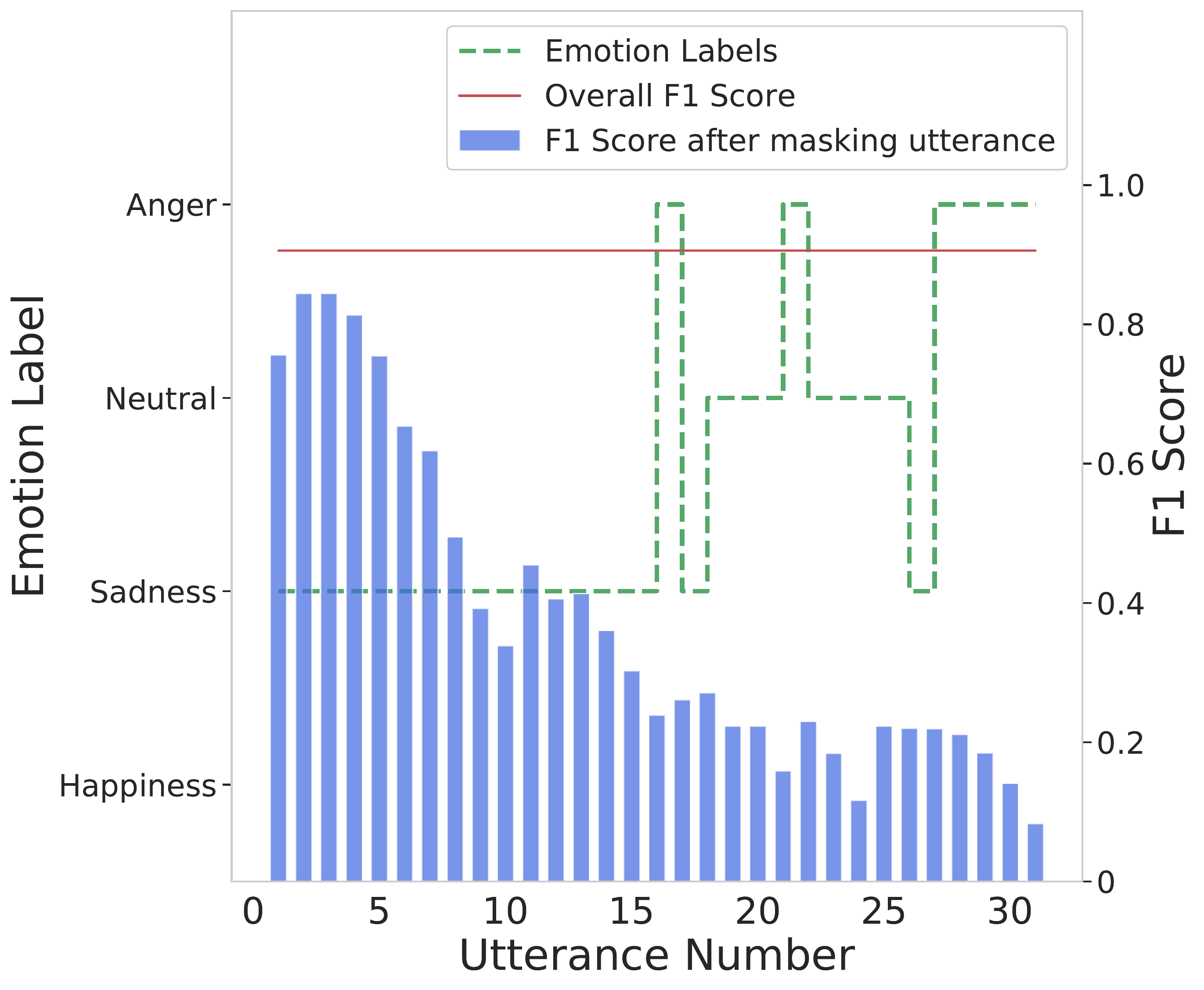}
  \caption{Importance of utterances in IEMOCAP classification.}
  \label{app:fig:importance-of-utterances-a3}
\end{figure}

\begin{figure*}[h]
\centering
  \includegraphics[width=\textwidth]{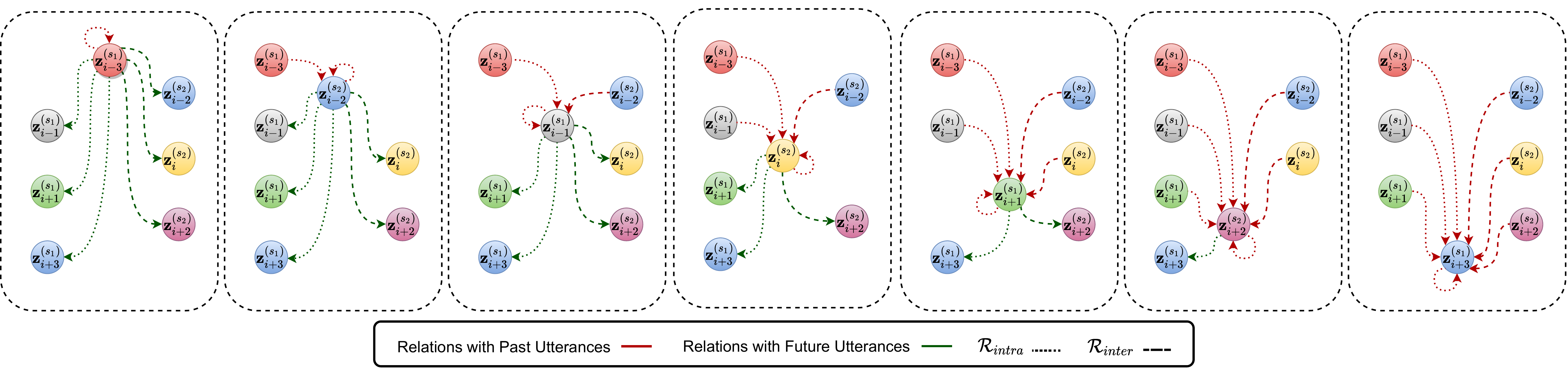}
  \caption{{Graph formation process in (\modelname) architecture. }}
  \label{fig:graph_formation}
\end{figure*}

\begin{table*}[hbt]
\renewcommand{\arraystretch}{1.6}
\centering
\setlength\tabcolsep{15pt}
\hspace{0.2cm}
\centering
\resizebox{\textwidth}{!}{
\begin{tabular}{|c|c|c|c|}
\hline
\textbf{Figure} & \textbf{Central Node}   & $\mathcal{R}_{intra}$ & $\mathcal{R}_{inter}$
\\ \hline
(a) &  $u^{(S_1)}_{i-3}$          & $u^{(S_1)}_{i-3} \leftarrow u^{(S_1)}_{i-3}, u^{(S_1)}_{i-3} \rightarrow u^{(S_1)}_{i-1}, u^{(S_1)}_{i-3} \rightarrow u^{(S_1)}_{i+1}, u^{(S_1)}_{i-3} \rightarrow u^{(S_1)}_{i+3}$             &  $u^{(S_1)}_{i-3} \rightarrow u^{(S_2)}_{i-2}, u^{(S_1)}_{i-3} \rightarrow u^{(S_2)}_{i}, u^{(S_1)}_{i-3} \rightarrow u^{(S_2)}_{i+2}$                   \\ \hline
(b)  & $u^{(S_2)}_{i-2}$         & $u^{(S_2)}_{i-2} \leftarrow u^{(S_2)}_{i-2}, u^{(S_2)}_{i-2} \rightarrow u^{(S_2)}_{i}, u^{(S_2)}_{i-2} \rightarrow u^{(S_2)}_{i+2}$             &  $u^{(S_2)}_{i-2} \leftarrow u^{(S_1)}_{i-3}, u^{(S_2)}_{i-2} \rightarrow u^{(S_1)}_{i-1}, u^{(S_2)}_{i-2} \rightarrow u^{(S_1)}_{i+1}, u^{(S_2)}_{i-2} \rightarrow u^{(S_1)}_{i+3}$                    \\ \hline
(c)   &     $u^{(S_1)}_{i-1}$    & $u^{(S_1)}_{i-1} \leftarrow u^{(S_1)}_{i-3}, u^{(S_1)}_{i-1} \leftarrow u^{(S_1)}_{i-1}, u^{(S_1)}_{i-1} \rightarrow u^{(S_1)}_{i+1}, u^{(S_1)}_{i-1} \rightarrow u^{(S_1)}_{i+3}$             &  $u^{(S_1)}_{i-1} \leftarrow u^{(S_2)}_{i-2}, u^{(S_1)}_{i-1} \rightarrow u^{(S_2)}_{i}, u^{(S_1)}_{i-1} \rightarrow u^{(S_2)}_{i+2}$                   \\ \hline
(d)    & $u^{(S_2)}_{i}$       & $u^{(S_2)}_{i} \leftarrow u^{(S_2)}_{i-2}, u^{(S_2)}_{i} \leftarrow u^{(S_2)}_{i}, u^{(S_2)}_{i} \rightarrow u^{(S_2)}_{i+2}$             &  $u^{(S_2)}_{i} \leftarrow u^{(S_1)}_{i-3}, u^{(S_2)}_{i} \leftarrow u^{(S_1)}_{i-1}, u^{(S_2)}_{i} \rightarrow u^{(S_1)}_{i+1}, u^{(S_2)}_{i} \rightarrow u^{(S_1)}_{i+3}$                    \\ \hline
(e)   & $u^{(S_1)}_{i+1}$       & $u^{(S_1)}_{i+1} \leftarrow u^{(S_1)}_{i-3}, u^{(S_1)}_{i+1} \leftarrow u^{(S_1)}_{i-1}, u^{(S_1)}_{i+1} \leftarrow u^{(S_1)}_{i+1}, u^{(S_1)}_{i+1} \rightarrow u^{(S_1)}_{i+3}$             &  $u^{(S_1)}_{i+1} \leftarrow u^{(S_2)}_{i-2}, u^{(S_1)}_{i+1} \leftarrow u^{(S_2)}_{i}, u^{(S_1)}_{i+1} \rightarrow u^{(S_2)}_{i+2}$                   \\ \hline
(f)   & $u^{(S_2)}_{i+2}$       & $u^{(S_2)}_{i+2} \leftarrow u^{(S_2)}_{i-2}, u^{(S_2)}_{i+2} \leftarrow u^{(S_2)}_{i}, u^{(S_2)}_{i+2} \leftarrow u^{(S_2)}_{i+2}$             &  $u^{(S_2)}_{i+2} \leftarrow u^{(S_1)}_{i-3}, u^{(S_2)}_{i+2} \leftarrow u^{(S_1)}_{i-1}, u^{(S_2)}_{i+2} \leftarrow u^{(S_1)}_{i+1}, u^{(S_2)}_{i+2} \rightarrow u^{(S_1)}_{i+3}$                   \\ \hline
(g)   & $u^{(S_1)}_{i+3}$       & $u^{(S_1)}_{i+3} \leftarrow u^{(S_1)}_{i-3}, u^{(S_1)}_{i+3} \leftarrow u^{(S_1)}_{i-1}, u^{(S_1)}_{i+3} \leftarrow u^{(S_1)}_{i+1}, u^{(S_1)}_{i+3} \leftarrow u^{(S_1)}_{i+3}$             &  $u^{(S_1)}_{i+3} \leftarrow u^{(S_2)}_{i-2}, u^{(S_1)}_{i+3} \leftarrow u^{(S_2)}_{i}, u^{(S_1)}_{i+3} \leftarrow u^{(S_2)}_{i+2}$                   \\ \hline
\end{tabular}}
\caption{Relations for each instance of Figure \ref{fig:graph_formation}, where relations with past utterances are denoted by ($\leftarrow$) and relations with future utterances are denoted by ($\rightarrow$)}
\label{app:tab:relations-for-instance}
\end{table*}

\begin{table}[!hbt]
\renewcommand{\arraystretch}{1.4}
\setlength\tabcolsep{11pt}
\hspace{0.2cm}
\centering
\resizebox{\columnwidth}{!}{
\begin{tabular}{|c|c|c|c|c|}
\hline
\textbf{Relation Type} & \textbf{Node A} & \textbf{Node B} & \textbf{Relation Causality}    & \textbf{Relation }
\\ \hline
1          & ${u^{(S_1)}}$ & ${u^{(S_1)}}$& Past& ${u^{(S_1)} \leftarrow u^{(S_1)}}$                           \\ \hline
2           & ${u^{(S_1)}}$ & ${u^{(S_2)}}$&  Past&${u^{(S_1)} \leftarrow u^{(S_2)}}$                        \\ \hline
3           & ${u^{(S_2)}}$ & ${u^{(S_1)}}$&  Past&${u^{(S_2)} \leftarrow u^{(S_1)}}$                             \\ \hline
4          & ${u^{(S_2)}}$ & ${u^{(S_2)}}$&  Past& ${u^{(S_2)} \leftarrow u^{(S_2)}}$                             \\ \hline
5          & ${u^{(S_1)}}$& ${u^{(S_1)}}$ &Future & ${u^{(S_1)} \rightarrow u^{(S_1)}}$                            \\ \hline
6          & ${u^{(S_1)}}$ & ${u^{(S_2)}}$&Future & ${u^{(S_1)} \rightarrow u^{(S_2)}}$                            \\ \hline
7     & ${u^{(S_2)}}$     & ${u^{(S_1)}}$ &Future & ${u^{(S_2)} \rightarrow u^{(S_1)}}$                                \\ \hline
8          & ${u^{(S_2)}}$& ${u^{(S_2)}}$ &Future & ${u^{(S_2)} \rightarrow u^{(S_2)}}$                                \\ \hline
\end{tabular}}
\caption{Unique Relation types for a conversation between two speakers}
\label{app:tab:unique-Relations}
\end{table}

\section{Additional Analysis} \label{app:addi-analysis}

\noindent\textbf {Efficacy of the GNN Layer:} We observe the efficacy of the GNN component in our architecture and visualize the features before GNN and after the GNN component (Figure \ref{app-fig:bef-aft-gnn6}) explained in section \ref{sec:result_analysis}. 


\noindent\textbf {Importance of utterances:} Figure \ref{app:fig:importance-of-utterances-a3} shows the obtained results for a dialogue instance taken randomly from IEMOCAP 4-way. For the first 15 utterances, emotions state being sadness, the effect of masking the utterances is more negligible for the first 5 utterances. This drop depicts the importance of utterances 5-15 that affect future utterances. Further, masking the utterances with high emotion shift (15 to 30) drops the F1 score of the dialogue, showing the importance of fluctuations for predicting the emotion states for other utterances.

\section{Graph Formation} \label{app:graph-Formation}
To give a clear picture of the graph formation procedure, we describe the process for utterances spoken in a dialogue. As an illustration, let's consider two speakers, $S_1$ and $S_2$, present in a conversation of 7 utterances. Features corresponding to each utterance is shown as a node in Figure \ref{fig:graph_formation}. Speaker 1 speaks utterances $u_{i-3}, u_{i-1}, u_{i+1}, u_{i+3}$ and Speaker 2 speaks $u_{i-2}, u_{i}, u_{i+2}$. After creating the graphs with relations, the constructed graph would look like shown in Figure \ref{fig:graph_formation}, and the corresponding relations for each instance would be as shown in Table \ref{app:tab:relations-for-instance}. Since there are two speakers in the conversation ($S_N=2$), the total number of unique relations would be:
\begin{align*}
    \text{number of relations} &= 2 \times (S_N)^2\\
    &= 2 \times (2)^2\\
    &= 8
\end{align*}

Table \ref{app:tab:unique-Relations} shows the number of possible unique relations for a conversation between two speakers.

\section{Discussion} \label{app:discussion}

\subsection{Modality Fusing Mechanisms}\label{app:fusion}
While experimenting with the model architecture, we explored various mechanisms for mixing information from multiple modalities. Some of the mechanisms include pairwise attention inspired from \citet{ghosal-etal-2018-contextual_pairwiseAttention}, bimodal attention present in Multilogue-Net \cite{shenoy2020multilogue}, and crossAttention layer proposed in HKT \cite{Hasan2021HumorKE_HKT}. However, in our case, none of these fusing mechanisms shows significant performance improvement over simple concatenation. Moreover, all these fusing mechanisms require extra computation steps for fusing information. In contrast, a simple concatenation of modality features works well with no additional computational overhead. 

\subsection{Effect of window size in Graph Formation} \label{app:windowSize}
To explore the effect of window size in the Graph Formation module of our architecture, we conduct experiments with multiple window sizes. The obtained results are present in Table \ref{tab:graph_window_size_iemocap_4}. The window size can be treated as a hyperparameter that could be adjusted while training our architecture. Moreover, the freedom of setting the window size makes our architecture more flexible in terms of usage. A larger window size would result in better performance for cases where the inter and intra speaker dependencies are maintained for longer sequences. In contrast, setting a lower window size would be better in a use case where the topic frequently changes in dialogues and speakers are less affected by another speaker. In the future, we plan to explore a dynamic and automatic selection of window size depending on the dialogue instance. 
\begin{table}[!h]\small
\renewcommand{\arraystretch}{1.5}
\setlength\tabcolsep{12pt}
\resizebox{\columnwidth}{!}{
\centering
\begin{tabular}{|c|c|c|c|}
\hline
\textbf{Modalities} & \textbf{Window Past} & \textbf{Window future} & \textbf{F1 Score (\%)} \\ \hline
atv                 & 1                    & 1                      & 81.72                      \\ \hline
atv                 & 2                    & 2                      & 83.21                      \\ \hline
atv                 & 4                    & 4                      & \textbf{84.08}                      \\ \hline
atv                 & 5                    & 5                      & 83.19                      \\ \hline
atv                 & 6                    & 6                      & 82.49                      \\ \hline
atv                 & 7                    & 7                      & 82.28                      \\ \hline
atv                 & 9                    & 9                      & 82.77                       \\ \hline
atv                 & 10                   & 10                     & \textbf{84.50}                      \\ \hline
atv                 & 11                   & 11                     & 83.93                       \\ \hline
atv                 & 15                   & 15                     & 83.78                      \\ \hline
\end{tabular}}
\caption{Results for various window sizes for graph formation on the IEMOCAP (4-way) dataset.}
\label{tab:graph_window_size_iemocap_4}
\end{table}